\documentclass[10pt,twocolumn]{article} 
\usepackage{simpleConference}
\usepackage{times}
\usepackage{graphicx}
\usepackage{algpseudocode,algorithm,algorithmicx}
\usepackage{setspace}
\usepackage{threeparttable}

\def\BibTeX{{\rm B\kern-.05em{\sc i\kern-.025em b}\kern-.08em
    T\kern-.1667em\lower.7ex\hbox{E}\kern-.125emX}}
\usepackage{balance}
\usepackage{array}
\newcolumntype{P}[1]{>{\centering\arraybackslash}p{#1}}
\newcolumntype{M}[1]{>{\centering\arraybackslash}m{#1}}
\newcolumntype{C}{>{\centering\arraybackslash}m{0.25\textwidth}}

\usepackage{amsmath,amssymb} 
\usepackage{color}
\usepackage{siunitx}
\usepackage[table]{xcolor}
\usepackage{caption}
\usepackage{epsfig} 
\usepackage{array}
\usepackage{booktabs}
\usepackage{multirow}

\begin{document}

\title{\fontsize{20}{20}\selectfont{CRC-RL: A Novel Visual Feature Representation Architecture for Unsupervised Reinforcement Learning}}

\author{Darshita Jain$^1$, Anima Majumder$^1$, Samrat Dutta$1$, Swagat Kumar$^2$
 \\
\\
$^{1}$ TATA Consultancy Services, Bangalore, India. \\
$^{2}$ Edge Hill University, UK. \\
\today
\\
\\
 $^1$(darshita.jain, anima.majumder, d.samrat)@tcs.com , $^{2}$ kumars@edgehill.ac.uk \\

}

\maketitle
\thispagestyle{empty}

\begin{abstract}
 This paper addresses the problem of visual feature representation
 learning with an aim to improve the performance of end-to-end
 reinforcement learning (RL) models. Specifically, a novel
 architecture is proposed that uses a heterogeneous loss function,
 called CRC loss, to learn improved visual features which can then be
 used for policy learning in RL. The CRC-loss function is a
 combination of three individual loss functions, namely, contrastive,
 reconstruction and consistency loss. The feature representation is
 learned in parallel to the policy learning while sharing the weight
 updates through a Siamese Twin encoder model. This encoder model is
 augmented with a decoder network and a feature projection network to
 facilitate computation of the above loss components. Through
 empirical analysis involving latent feature visualization, an attempt
 is made to provide an insight into the role played by this loss
 function in learning new action-dependent features and how they are
 linked to the complexity of the problems being solved. The proposed
 architecture, called CRC-RL, is shown to outperform the
 existing state-of-the-art methods on the challenging Deep mind control suite
 environments by a significant margin thereby creating a new benchmark
 in this field. 

\end{abstract}
\label{sec:introduction}
In the recent past, deep reinforcement learning (DRL) algorithms have been
successfully used to learn action policies directly from visual
observations, thereby finding application in several interesting areas such as gaming
\cite{mnih2015human, xue2022event}, robotics \cite{levine2016end, qureshi2018intrinsically, wang2021modular, nakamura2007reinforcement} and
autonomous vehicles \cite{yang2018visual, zhu2017target} etc. This success
is mostly driven by the agent's ability to jointly learn feature
representation and policy decisions by using long-term
credit-assignment capabilities of reinforcement learning algorithms in
an end-to-end fashion.  In spite of this success, RL algorithms are
known to be sample-inefficient and suffer from poor generalization for
high-dimensional observations such as images
\cite{laskin2020reinforcement, laskin2020curl}. There are several
approaches to address these concerns, including methods such as, transfer learning
\cite{zhuang2020comprehensive, islam2022transfer}, meta-learning \cite{jaafra2018review}
\cite{hospedales2021meta} and active learning
\cite{settles2009active}. \emph{Feature representation learning}
\cite{bengio2013representation} is an alternative, and sometimes
complementary, to these approaches which aims at learning useful
features that can simplify the intended task, e.g., classification or
prediction. This paper primarily focuses on this later approach as it is now widely accepted that the problem of
sample-inefficiency in RL can be solved to a great extent by learning
suitable feature representation which is shown to expedite the policy
learning process \cite{yarats2021improving}.  The feature representations are
learned using self-supervised methods which are increasingly becoming popular
as they obviate the need for manually generated labeled datasets
thereby simplifying the practical deployment of deep learning models
\cite{ericsson2022self}. Some of these approaches include
auto-encoders \cite{bank2020autoencoders}, GANs \cite{lin2017marta}
\cite{peng2019cm}, contrastive learning \cite{chen2020simple} and data
augmentation techniques \cite{laskin2020reinforcement}
\cite{hansen2021generalization}. The features thus obtained have been
shown to greatly improve the sample efficiency and generalizability of
RL methods as demonstrated in \cite{finn2015learning}
\cite{laskin2020curl} \cite{hansen2021generalization}. 

Rather than decoupling the representation learning from policy
learning as done in  \cite{stooke2021decoupling}
\cite{hansen2021generalization}  \cite{yarats2021improving}, we
continue working with end-to-end models because of
their simplicity and aim at improving their performance by performing
auxiliary tasks as demonstrated in \cite{laskin2020curl}
\cite{jaderberg2016reinforcement}. Since the feature representations
are learned along side the policy decisions in an end-to-end fashion,
the features learned are actually \emph{action-dependent}. This is
because, the backward gradient flow from the policy-learning algorithm
is allowed to update the encoder weights. This makes the learned
feature vectors strongly correlated to the actions being
taken by the agent \cite{xenou2018deep}. Given this hindsight, we are
motivated by two factors. First, it is our belief that the
quality of the features learnt could be improved by using a better loss
function leading to improved RL performance.
Secondly, we are keen to develop a better understanding of  
the relationship between the feature and action spaces.  Towards
fulfilling these objectives, we propose
a new heterogeneous loss function called \emph{CRC loss} for feature
representation learning by combining three different loss functions,
namely, \emph{contrastive loss} \cite{laskin2020curl},
\emph{reconstruction loss} and \emph{consistency loss}.   The
reconstruction loss obtained with an auto-encoder model helps in
learning compact features that is sufficient to reconstruct the original
observations. On the other hand, the consistency loss
\cite{hansen2021generalization} helps in learning features that are
invariant to image augmentations. In other words, by minimizing
the consistency loss, the encoder is encouraged to learn
\emph{task-relevant}
features while ignoring irrelevant aspects (such as, background color) thereby avoiding
observational over-fitting
\cite{neyshabur2020observational}.  Similarly, the contrastive
loss helps in learning \emph{class-invariant} features from augmented images by
contrasting them against a batch of negative samples. In that sense,
these three loss functions contribute complementary information and
hence, should improve the feature representation learning when
combined together. In order to
implement feature training with this loss function, a new architecture
inspired by CURL \cite{laskin2020curl}, is proposed that uses a
Siamese Twin encoder model, a decoder network and a feature predictor
to generate these losses. Through empirical analysis including feature
visualizations, it is shown that the feature representations learnt by the CRC
loss function is different from those learnt with the baseline CURL
model, indicating the role played by the CRC-loss in learning new
action-dependent features. In addition, visualization of correlation
matrices between latent features generated by this model show increasingly
complex patterns for complex environments with higher-dimensional action
spaces, thereby providing a clue about how features are inherently
linked with action in an end-to-end RL model. Through rigorous
experiments on the challenging Deep Mind Control suite environments
\cite{tassa2018deepmind}, it is shown that the proposed CRC-RL model
outperforms the existing state-of-the-art methods  by a significant
margin, thereby establishing the efficacy of the approach. The design
choices for the proposed model are justified through extensive
ablation studies. 

In short, the major contributions made in this paper are as follows:
\begin{enumerate}
    \item A new self-supervised feature representation architecture
      along with a novel loss function is proposed to improve the
      performance of RL models in learning policies directly from
      image observations.  
    \item Through empirical analysis involving latent feature
      visualization, an attempt has been made to provide insights into
      the relationship between the action and feature space thereby
      providing better standing of the role played of the new loss
      function in learning \emph{action-dependent} features.  
    \item The resulting architecture is shown to outperform existing
      state-of-the-art methods on the challenging DMC environments by
      a significant margin. 
  \end{enumerate}
  The rest of this paper is organized as follows. Related works are
  reviewed in the next section. The proposed architecture is described
  in Section \ref{sec:prop}. The experimental results are discussed
  and analyzed in Section \ref{sec:exp}. The paper ends with the
  concluding section \ref{sec:conc}. 
 	\begin{figure*}[t]
		\centering
		\includegraphics[scale=0.3]{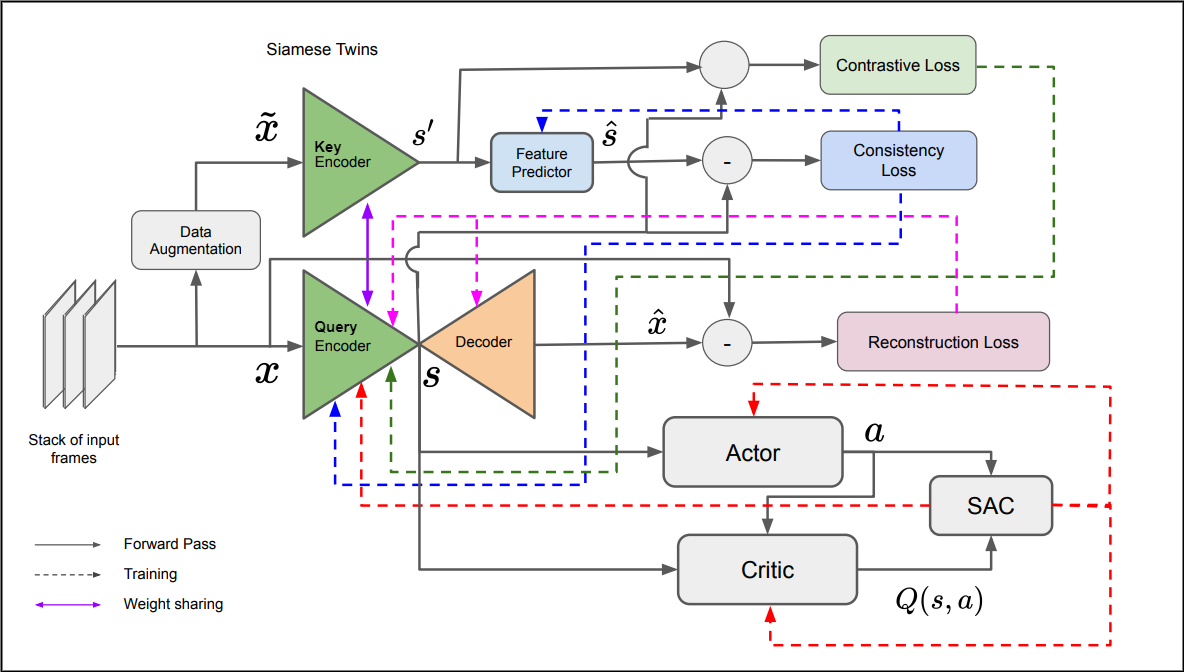}
      \caption{\footnotesize{CRC-RL Architecture: It consists of a
        Siamese Twin encoder along with a decoder and a feature
        predictor network. The query encoder together with the decoder
        forms an auto-encoder. The query encoder is used for learning
        policy using SAC algorithm.  Observations are  data-augmented
        to form query and key observations, which are then encoded
        into latent features by the respective encoders.   Only the
      query encoder weights are updated during the training step. The
    weights of key encoder are exponential moving average of query
  encoder weights.}}
		\label{fig:arch} 
	\end{figure*}

\section{Related Works}
\label{sec:related}
This section provides an overview of related literature in the following subsections. 

\subsection{Deep RL architectures for policy learning}
Reinforcement learning algorithms  learn the optimal policy for a
given task by maximizing a measure of \emph{cumulative discounted future reward} for
the task while balancing between \emph{exploration} (of new
possibilities) and \emph{exploitation} (of post
experiences) \cite{sutton2018reinforcement}. This cumulative discounted
reward function, represented as Q or \emph{value} function, is not known a
priori and, is used to evaluate a given action taken by the agent.
Depending on how this function is estimated and desirable actions are
derived from it, the RL-based methods can be broadly classified into
two categories: \emph{value-based} methods and \emph{policy-based
methods}. The value-based methods aim at estimating the Q-function
and then derive action from this by using a greedy policy. On the other hand, policy-
based methods directly estimate the policy function by maximizing a given objective
function. The traditional Q-learning
algorithm  estimates the Q function iteratively by using an
approximate dynamic programming formulation based on Bellman's
equation starting from an initial estimate
\cite{barto1995reinforcement}. The original Q-learning algorithm could
be applied to problems with discrete state (observation) and action
spaces and hence, suffered from the \emph{curse-of-dimensionality} problem
with higher dimensions and range of values. This limitation is
overcome by using a deep network to estimate Q function that can take
arbitrary observation inputs, thereby, greatly enhancing the capabilities of RL
algorithms. The resulting approach is known as Deep Q Networks (DQN)
\cite{van2016deep} \cite{sewak2019deep} which has been applied
successfully to a wide range of problems while achieving superhuman
level performances in a few cases, such as ATARI video games
\cite{mnih2013playing}, Go \cite{holcomb2018overview} etc.  The
success of DQN has spawned a new research field known as deep
reinforcement learning (DRL) attracting a large following of
researchers. Readers are referred to \cite{arulkumaran2017deep} for a
survey of this field. The DQN models were subsequently extended to
continuous action spaces by using policy gradient methods that used a
parameterized policy function to maximize DQN output using gradient
ascent methods \cite{lillicrap2015continuous}
\cite{duan2016benchmarking}. This has opened the doors for solving
various robotics problems that use continuous values such as joint
angles, joint velocities or motor torques as input. Since then, a
number of methods have been proposed to improve the performance of RL
algorithms and have been applied successfully to different robotic
problems - manipulation \cite{gu2017deep} \cite{nguyen2019review},
grasping \cite{quillen2018deep} \cite{joshi2020robotic}, navigation
\cite{yue2019experimental} etc. Some of the notable methods include
actor-critic models - A2C and A3C
\cite{mnih2016asynchronous}, soft actor-critic (SAC)
\cite{haarnoja2018soft} and proximal policy optimization (PPO)
\cite{schulman2017proximal}. In spite of the success of these
methods, (deep) reinforcement learning algorithms, in general, suffer
from limitations such as poor sampling efficiency leading to longer
training time, poor generalization and instability. The work presented
in this paper aims to address some of these concerns by focusing on
learning better task-relevant features.

\subsection{Feature Representation Learning in RL}

It is now widely accepted that learning policies from low-dimensional
feature vectors based on physical states is much more sample efficient
compared to learning directly from image pixels \cite{laskin2020curl}
\cite{botteghi2022unsupervised}. Hence, it is
imperative to learn suitable state representations from image
observations that will reduce the search space thereby improving the
sample efficiency and stability of RL algorithms. The field of
self-supervised representation learning has seen great progress in
last few years. Auto-encoders \cite{bank2020autoencoders}
\cite{pinaya2020autoencoders} learn the state representation by
compressing the observation into low-dimensional state that is
sufficient to reconstruct the observation. These have been used to
improve the performance of RL algorithms as demonstrated in
\cite{yarats2021improving}\cite{raffin2018s} \cite{lange2010deep}
\cite{finn2015learning}.  On the other hand, contrastive learning
\cite{chen2020simple} \cite{kolesnikov2019revisiting} learns the
class-relevant feature representations by maximizing the agreement between the
augmented versions of the same observation. It has been shown to
greatly improve the sample efficiency of RL algorithms as in
\cite{laskin2020curl}. Similarly, recent studies have shown that the right
kind of data augmentation techniques can improve the sample efficiency
and generalization capabilities of RL algorithms learning
\emph{task-relevant} features which remain unaffected by distractions
introduced by the augmentation \cite{laskin2020reinforcement}
\cite{raileanu2020automatic}. This can be further enforced by making
the encoder minimize the consistency loss as suggested in
\cite{hansen2021generalization}. In short, learning suitable feature
representation plays a significant role in improving the performance
of RL algorithms by increasing sample efficiency, improving
generalization and stability. The work presented in this paper
contributes to this field by proposing a novel loss function that
leads to superior learning performance for continuous control tasks as
will be demonstrated later in this paper.

\section{Method} \label{sec:prop}
This section provides details of the proposed CRC-RL model that
uses a novel heterogeneous loss function to extract useful information
from visual images to be used for learning optimal policy using an
end-to-end RL framework. The discussion is organized in the following
subsections. The architecture of the proposed model is
described next.   


\subsection{The Model Architecture}
  The overall architecture of the proposed model is shown in Figure
  \ref{fig:arch}. The observation is available in the form of images
  which are stacked together to act as input to the model. Stacking of
  frames is a heuristic approach to incorporate temporal information
  in the learning process \cite{shang2021reinforcement}. The
  observations obtained from the environment is stored in a replay
  buffer $\mathcal{D}$ and a batch is sampled from this replay buffer
  during the training process. A Siamese
  Twin encoder model is employed for extracting features from the
  input images.  These two encoders, termed as \emph{query} and
  \emph{key} encoders, are used for computing contrastive and
  consistency losses. The query encoder with a decoder is used for
  computing the reconstruction loss. A combination of these three
  losses, known as the CRC loss, is used for updating the parameters of
  the query encoder and decoder network.  The input images are augmented
  before applying to the key encoder. The features obtained from the
  query encoder is used for policy estimation using soft-actor-critic
  algorithm \cite{haarnoja2018soft}. The parameters of the query
  encoder and decoder networks are updated using error signals obtained
  from their own outputs as well as from the RL algorithm. Since the
  encoder networks are getting influenced by the RL policy algorithm,
  the features learnt in the process are \emph{action-dependent}. This
  aspect will be analyzed in more detail in the experiment
  section presented later in this paper. The weights of the key encoder network
  is the exponential moving average of the query encoder weights. The
  proposed CRC loss function used for learning the feature embeddings
  is discussed in the next subsection. 

%

  \subsection{The loss function for feature extraction}
  The query encoder is trained using the proposed CRC loss function
  which is a combination of the following three loss components as
  described below.

       \subsubsection{Contrastive loss}
      
    In contrastive learning, we have a query $q$ observation and a set
    of key observation samples $\mathbb{K} = \{k_0, k_1,...\}$
    consisting of both positive samples $(k_+)$ and     negative
    samples $(\mathbb{K} \setminus {k_+})$. The positive samples are
    those that belong to the same class as that of the query
    observation and the rest are considered to be the negative
    samples. The goal is to learn embeddings such that $q$ is
    relatively more similar     to the positive keys $k_+$ than the
    negative keys in the latent space. The query and key observations,
    generated by applying data augmentation on sampled observations,
    are encoded using the query and key encoder respectively.    The
    contrastive loss depends on the output of both the encoders
    (Siamese Twin) represented by the symbols $\mathbf{s}$ and
    $\mathbf{s}'$ respectively.  The
    idea behind using the contrastive loss is that the different
    augmentations of the same image will have the same underlying
    information and hence their high-level representations will be
    mapped together in the latent space. The similarity between the
    query and the key embeddings is computed using the bilinear
    inner-product $q^TWk>0$ where $W$ is a symmetric matrix of
    parameters to be estimated 
    \cite{henaff2020data} along with other parameters during the
    training process. The objective of training is to reduce this
    similarity measure so that the query embeddings become more
    distinct from the key embeddings over time ($q^TWk \approx 0
    \Rightarrow q\perp k,\; W > 0$). This is achieved by minimizing the 
    InfoNCE loss \cite{oord2018representation} given by:

    \begin{equation}
    L_q = \log\frac{\exp(q^TWk_+)}{\sum_{k_i \in \mathbb{K}}\exp(q^TWk_i)}
    \end{equation}
    
    \subsubsection{Reconstruction loss}
    A well-trained encoder-decoder network is expected to reconstruct
    the input image at the output of the decoder network. The
    reconstruction loss is computed based on the inaccuracy in the
    reconstructed image. A convolutional encoder $f_\theta$ maps an
    input observation $\mathbf{x}\in \mathbb{R}^{m\times n \times 3}$
    to a lower-dimensional latent vector $\mathbf{s}\in \mathbb{R}^l$,
    and a deconvolutional decoder $g_\phi$ then reconstructs
    $\mathbf{s}$ back to $\mathbf{\hat{x}}\in \mathbb{R}^{m\times n
    \times 3}$ such that
    \begin{align}
    f_{\theta}: \mathbf{x} \rightarrow \mathbf{s} \\
    g_{\phi}: \mathbf{s} \rightarrow \mathbf{\hat{x}}
\end{align}     
    Both the encoders and decoder are trained simultaneously by
    maximizing the expected log-likelihood. The reconstruction loss
    checks how well the image has been reconstructed from the input.
    The reconstruction loss forces the update such that the latent
    representation preserves the core attributes of the input data.
    An $L_2$ penalty is imposed on the learned representation
    $\mathbf{s}$ and a weight-decay is imposed on the decoder parameters to
    incorporate the regularization affects as proposed in
    \cite{ghosh2019variational}. 
    \begin{equation}
    L_r = \mathbb{E}_{\mathbf{x} \sim D}[\log p_\theta(\mathbf{x}|\mathbf{s}) + \lambda_s\| \mathbf{s} \|^2 + \lambda_\theta \| \theta \|^2]
    \end{equation}
    where  $\lambda_s$, and $\lambda_\theta$ are hyper-parameters.
 \subsubsection{Consistency loss}   
    The consistency loss depends on the  output of both the query and
    key encoder $f_\theta$ and $f'_\theta$. Here, the query encoder
    takes the original non-augmented observation $\mathbf{x}$ and the
    key encoder uses the augmented observation $\mathbf{\Tilde{x}}$ as
    input. The output of the Key encoder $\mathbf{s}'$ is then
    used as an input to a feature predictor module, which is nothing but an MLP, to
    estimate the non-augmented embedding $\mathbf{\hat{s}}$. The
    consistency loss is designed to minimize the error between the
    non-augmented embedding $\mathbf{s}$ and
    the augmented embedding $\mathbf{s'}$, thereby enabling the encoder to learn
    essential \emph{task-relevant} features while ignoring irrelevant
    distractions (such as background clutter or texture). 
    This eliminates the need of using negative
    samples for the computation of consistency loss. The consistency
    loss function can, therefore, be mathematically written as:
     \begin{equation}
       L_c(\mathbf{\hat{s}}, \mathbf{s}, \theta) =
      \mathbb{E}_{\mathbf{x} \sim \mathcal{D}}[\|\mathbf{\hat{s}} -
      \mathbf{s} \|^2] \end{equation} 
    \subsection{The CRC loss function}
    It is our conjecture that each of the above three loss functions enables the encoder to
    extract non-redundant and complementary information from the
    higher dimensional input. Thus, a combination of these three
    should improve the overall RL performance in learning optimum
    policy. The resulting loss function, called CRC loss, has the
    following mathematical form: 
    \begin{equation} L_{CRC} = c_1L_q + c_2L_r +
      c_3L_c \end{equation} 
    where $c_i>0, \sum_i c_i = 1, i=1,2,3$ are hyper-parameters that
    control the relative importance of individual components.  The RL
    model for policy learning takes the query encoder output as its
    input. The SAC algorithm used for learning policy is also allowed
    to affect the query encoder weights $f_\theta$ during the backward gradient
    update step. At regular
    intervals, the key encoder $f'_\theta$ weights are updated using
    the exponential moving average (EMA) of the weights of the query
    encoder $f_\theta$. The feature learning and policy learning takes
    place in jointly in parallel. The latent representations learned by the query
    encoder $f_\theta$ receives gradients from both the CRC loss and
    the SAC algorithm losses. This makes the feature representations
    \emph{action-dependent}, an aspect which will be analyzed in some
    more detail in the next section. 

\section{Experimental Results and Discussions}  \label{sec:exp} 

   The proposed CRC-RL model architecture takes its inspiration from
   the original CURL implementation by Laskin et al.
   \cite{laskin2020curl}. The original model is extended by
   incorporating additional decoder and feature predictor to
   facilitate computing the CRC loss function as described in the
   previous section. The model is implemented using PyTorch
   \cite{paszke2019pytorch} deep learning framework. The reinforcement learning framework for policy estimation makes use of the publicly released implementation of the SAC algorithm by Yarats
   et al. \cite{yarats2020image}. The query encoder and decoder
   architecture is similar to the ones used in the above work. The
   query encoder weights are tied between the actor and critic so that
   they both use the same encoder to embed input image observations.
   The feature predictor module is a MLP network which consists of
   cascaded linear layers and ReLU activation function. The complete
   list of hyper-parameters is shown in Table \ref{tab:hyper_params}.  A
   number of experiments are carried out to establish the efficacy of
   the proposed model. The design choices are justified through
   several ablation studies as discussed below.

   \begin{table}[!t] 
     \centering
      \captionsetup{width=0.60\linewidth}
     \caption{\footnotesize{Hyper-parameters used for DMControl
       experiments. Most hyper-parameters values are unchanged across
       environments with the exception for action repeat, learning
     rate, and batch size.}} \scalebox{1.0}{ \begin{tabular}{ |c|c| }
       \hline Hyper-parameter & Value \\ \hline Pre transform image
       size & (100, 100) \\ Image size & (84, 84) \\ Action repeat &
       8 \\ Frame stack & 3 \\ Transform & Random crop \\ Replay
       buffer capacity & 100000 \\ Initial steps & 1000 \\ Batch size
       & 512 \\ Hidden layers & 1024 \\ Evaluation episodes & 10 \\
       Optimizer & Adam \\ Learning rate $(f_{\theta}, \pi_{\psi},
       Q_{\phi})$ & 1e-3 \\ Learning rate $(\alpha)$ &  1e-4 \\
       Critic target update frequency & 2 \\ Convolution layers & 4
       \\ Number of filters & 32 \\ Latent dimension & 50 \\ Discount
       $(\gamma)$ & 0.99 \\ Initial temperature & 0.1 \\ \hline
       \label{tab:hyper_params}
     \end{tabular}}  
   \end{table}

\begin{table*}
  \centering
       \captionsetup{width=1.0\linewidth}
    \caption{\footnotesize{Mean episodic reward (with standard
      deviation) over 10 evaluation runs on DMControl
      environments after training for 100k
    environment steps. The best scores are shown in bold letters. }}
    \label{tab:perf_comp_100k}
\resizebox{\textwidth}{!}{
   \begin{tabular}{|c|c|c|c|c|c|c|c|c|c| } \hline 100K
  Step & Our Method & CURL \cite{laskin2020curl} & SODA
  \cite{hansen2021generalization} & PLANET \cite{hafner2019learning} &
  DREAMER \cite{hafner2019dream} & SAC+AE \cite{yarats2021improving} &
  PIXEL & STATE & \% Increase  \\ 
     
     Scores &  &  &  & &  &  &  SAC \cite{haarnoja2018soft} & SAC
     \cite{haarnoja2018soft} & over CURL \\
     
     \hline FINGER, SPIN & \textbf{793${\pm}$36} & 767${\pm}$56 &
     363${\pm}$185 & 136${\pm}$216 & 341${\pm}$70 & 740${\pm}$64 &
     179${\pm}$66 & 811${\pm}$46 & 3.38 \\ CARTPOLE, SWINGUP &
     \textbf{813${\pm}$45} & 582${\pm}$146 & 474${\pm}$143 &
     297${\pm}$39 & 326${\pm}$27 & 311${\pm}$11 & 419${\pm}$40 &
     835${\pm}$22 & 39.6 \\ REACHER, EASY & \textbf{636${\pm}$301} &
     538${\pm}$233 & - & 20${\pm}$50 & 314${\pm}$155 & 274${\pm}$14 &
     145${\pm}$30 & 746${\pm}$25 & 18.2  \\ CHEETAH, RUN &
     \textbf{355${\pm}$31} & 299${\pm}$48 & - & 138${\pm}$88 &
     235${\pm}$137 & 267${\pm}$24 &  197${\pm}$15 & 616${\pm}$18 &
     18.7 \\ WALKER, WALK & 490${\pm}$52 & 403${\pm}$24 &
     \textbf{635${\pm}$48} & 224${\pm}$48 & 277${\pm}$12 &
     394${\pm}$22 &  42${\pm}$12 & 891${\pm}$82 &  21.5\\ BALL IN CUP,
     CATCH & \textbf{832${\pm}$81} & 769${\pm}$43 & 539${\pm}$111 &
     0${\pm}$0 & 246${\pm}$174 & 391${\pm}$82 & 312${\pm}$63 &
     746${\pm}$91 &  8.19 \\ \hline \end{tabular}} \end{table*} 


\subsection{Performance Comparison}
The performance of the proposed CRC-RL model is compared with the
current state-of-the-art methods on the challenging Deep mind control
suite (DMControl) environments \cite{tassa2018deepmind}. The outcome
is shown in Table \ref{tab:perf_comp_100k} and
\ref{tab:perf_comp_500k} after training for 100K and 500K environment
steps respectively. It can be observed that the proposed CRC-RL model 
outperforms the current state-of-the-art methods, such as CURL
\cite{laskin2020curl}, SODA \cite{hansen2021generalization}, PlaNet
\cite{hafner2019learning}, Dreamer \cite{hafner2019dream}, SAC+AE
\cite{yarats2021improving}, pixel-based SAC \cite{haarnoja2018soft} on
most of the DMControl environments, thereby establishing the
superiority of our approach. The environments shown in Table
\ref{tab:perf_comp_500k} are difficult compared to those shown in
Table \ref{tab:perf_comp_100k} and hence require longer training time.
In this case, our proposed model outperforms the baseline CURL model
in only 300K training steps. 

\begin{table}[ht]
  \centering
  \captionsetup{width=1.0\linewidth}
  \caption{\footnotesize{Mean episodic score (with standard
    deviation) for 10 evaluation runs on DMControl environments
  obtained after training for 500k environment steps. The best scores
are shown in bold letters.}}
  \begin{threeparttable}
    \scalebox{0.70}{
      \begin{tabular}{ |c|c|c|c| } 
        \hline
        Environment & Our Method\tnote{*} & CURL & \% Increase over CURL \\ 
        \hline
        QUADRUPED, WALK & $\mathbf{88\pm51}$ & 39${\pm}$22 & 125.6\\ 
        HOPPER, HOP & $\mathbf{61\pm33}$ & 10${\pm}$17 & 510\\ 
        WALKER, RUN & $\mathbf{306\pm5}$ & 245${\pm}$32 & 24.8\\ 
        FINGER TURN, HARD & $\mathbf{423\pm78}$ & 207${\pm}$32 & 104.3\\ 
        \hline
      \end{tabular}}
      \begin{tablenotes} \footnotesize
      \item[*] shows values for 300K training steps
      \end{tablenotes}
    \end{threeparttable}
    \label{tab:perf_comp_500k}
  \end{table}

\subsection{t-SNE Visualizations}
\begin{figure*}
  \centering
  \begin{tabular}{M{1cm}CCC} 
    \rotatebox{90}{CURL} & \includegraphics[scale=0.3]{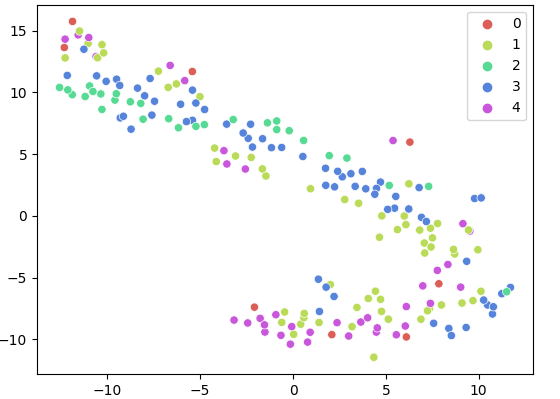} & 
    \includegraphics[scale=0.3]{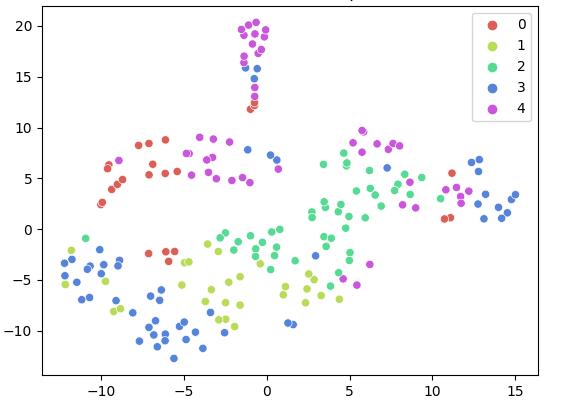} & 
    \includegraphics[scale=0.3]{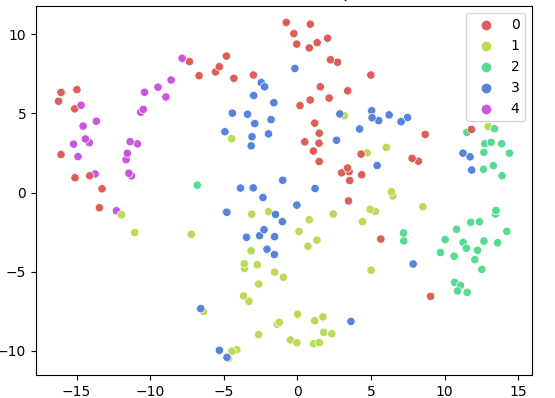} \\
    \rotatebox{90}{CRC-RL} & \includegraphics[scale=0.3]{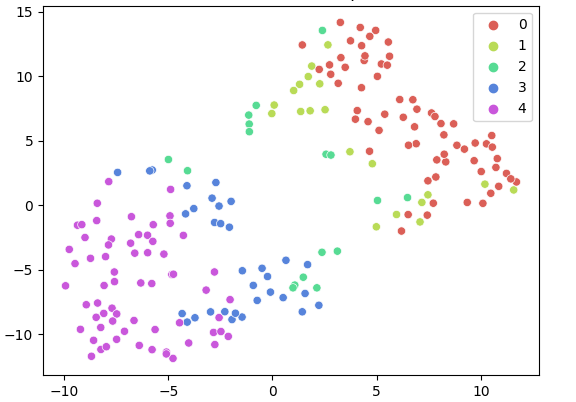} & 
    \includegraphics[scale=0.3]{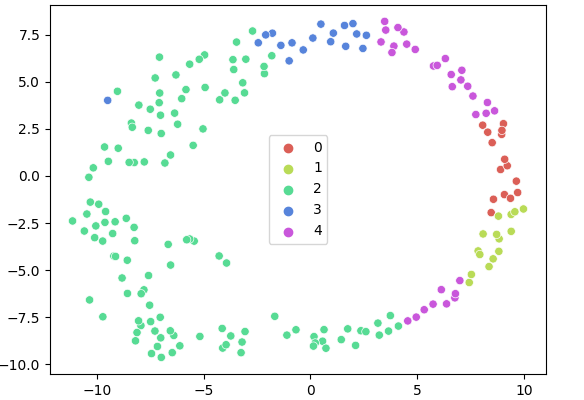} & 
    \includegraphics[scale=0.3]{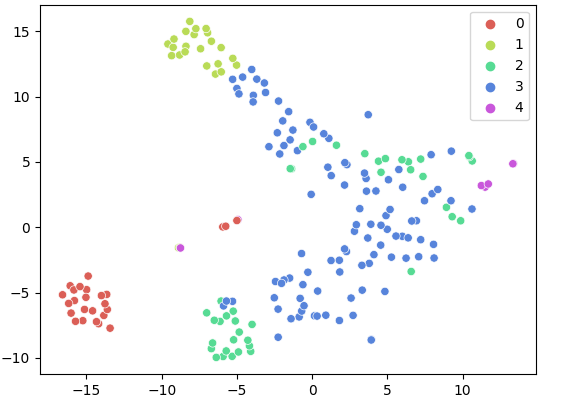} \\
    & \footnotesize{(a) Cartpole-Swingup} & \footnotesize{(b)
    Cheetah-Run} & \footnotesize{(c) Walker-walk}           
 \end{tabular}
  \caption{\footnotesize{t-SNE visualization of latent feature embeddings
  obtained from query encoder at 49K training steps. Colors correspond
to cluster labels in the action space. One can observe that CRC-RL
leads to more pristine clusters with less outliers compared to CURL.}}
  \label{fig:tsne_plot}
\end{figure*}

To better understand the relationship between the learned latent
representations and the action generated by the RL policy, we generate
the two-dimensional t-SNE plots of feature embeddings obtained from
the query encoder for 3 different environments as shown in Figure
\ref{fig:tsne_plot}. These features are assigned with the
corresponding action labels generated by partitioning the action space
into five clusters by using the k-mean clustering algorithm. As one
can observe, the proposed CRC-RL model leads to more pristine clusters
with lesser outliers compared to the baseline CURL
\cite{laskin2020curl} algorithm for some amount of training. Compared
to the Cartpole environment, other two are comparatively more complex
and require larger amount of time for training. This shows that the
proposed model leads to better correlation between the feature
embeddings and agent actions. This aspect has not been empirically
investigated extensively in the existing literature and thus, the
current work makes a novel contribution by filling this void. 

%

\begin{figure*}
\centering
\begin{tabular}{ccc}
\includegraphics[scale=0.07]{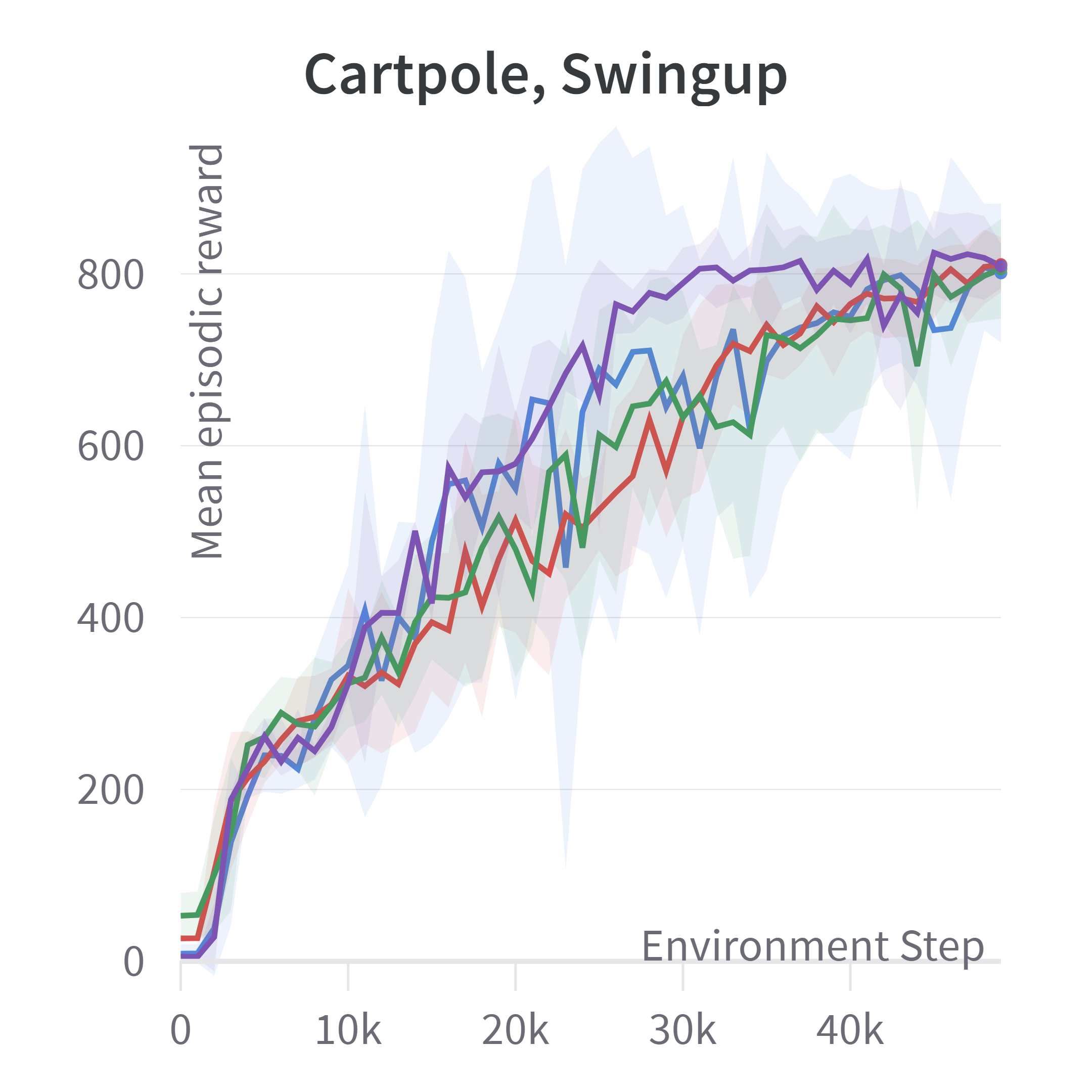}&
\includegraphics[scale=0.07]{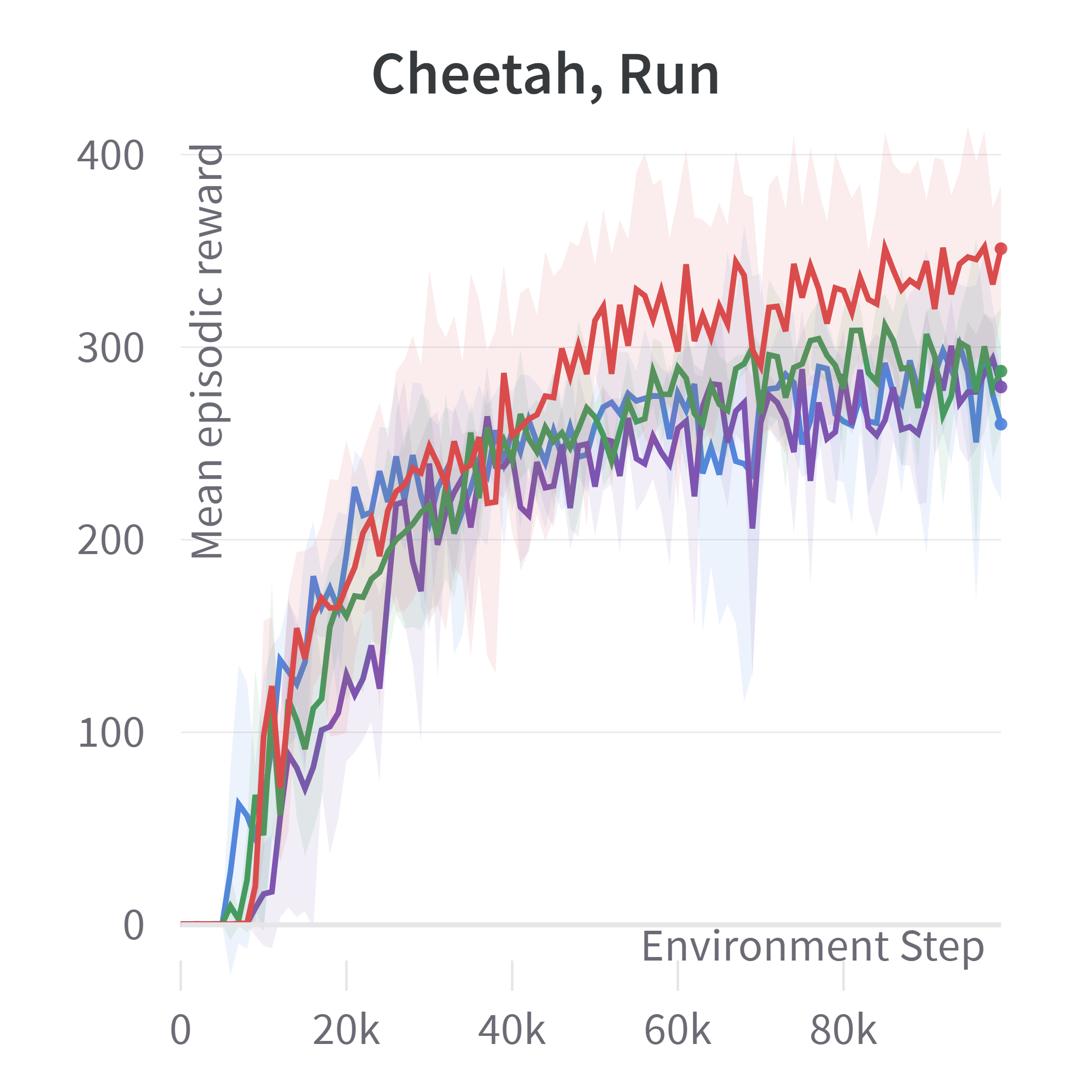}&
\includegraphics[scale=0.07]{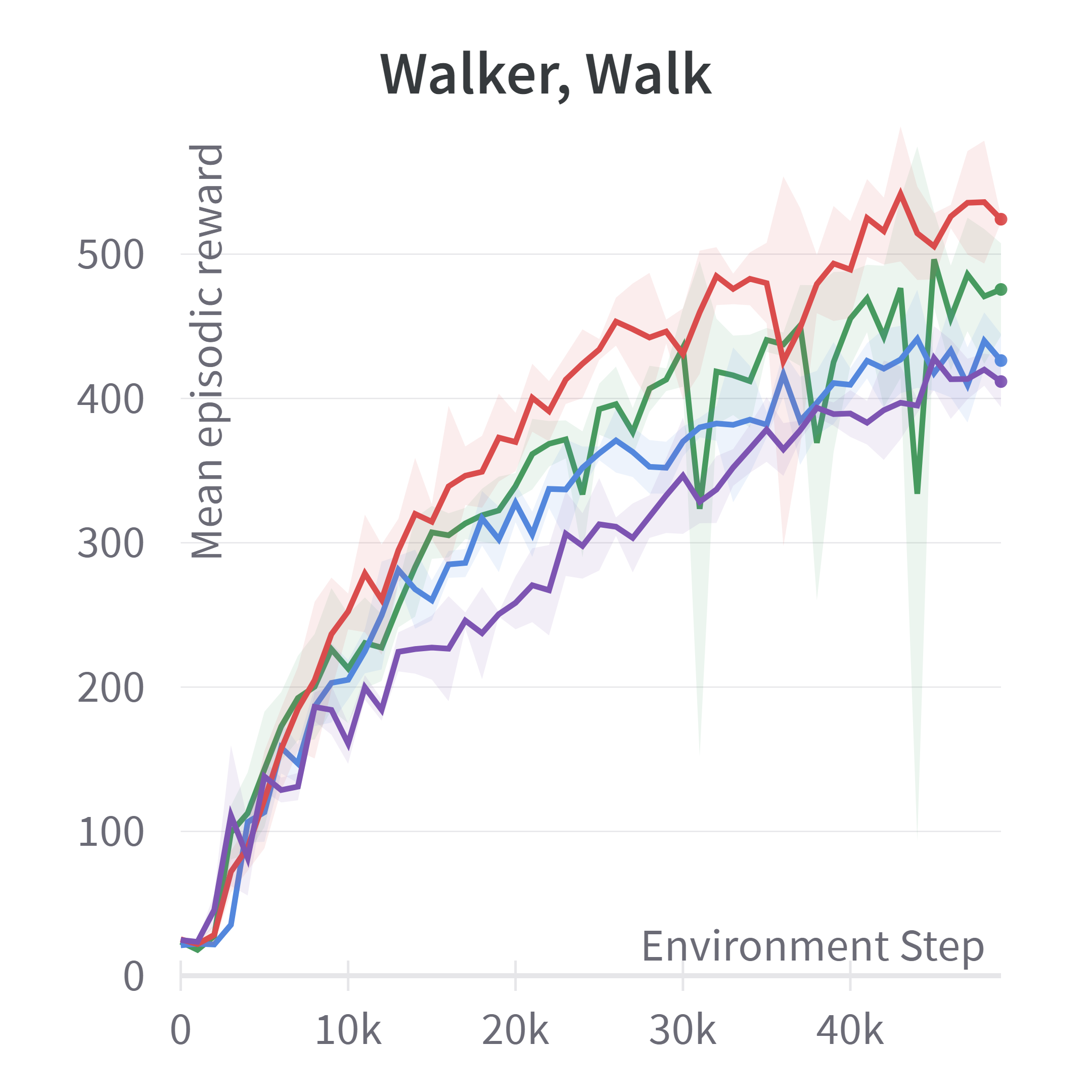}\\
\multicolumn{3}{c}{\includegraphics[scale=0.3]{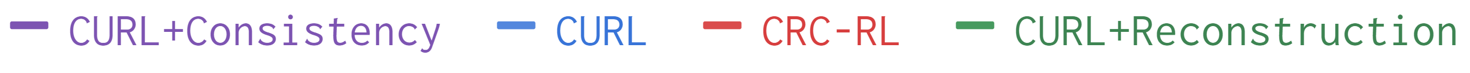}}
\\
\footnotesize{(a)} & \footnotesize{(b)} & \footnotesize{(c)}
\end{tabular}
\caption{\small{Effect of incorporating various loss components. CRC
  loss function performs better than other combinations for more
  difficult environments such as 'Cheetah-Run' and 'Walker-walk'. }}
  \label{tab:loss_functions}
\end{figure*}


\subsection{Feature Correlation Heat Maps}
\begin{figure*}[!t]
  \centering
  \begin{tabular}{CCC}
    \includegraphics[scale=0.4]{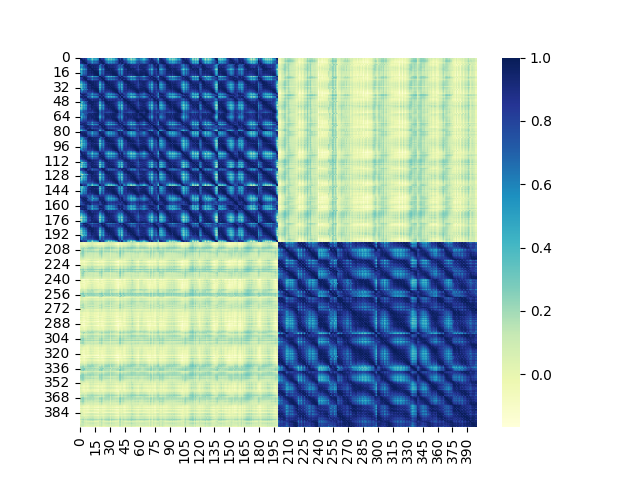} & 
    \includegraphics[scale=0.4]{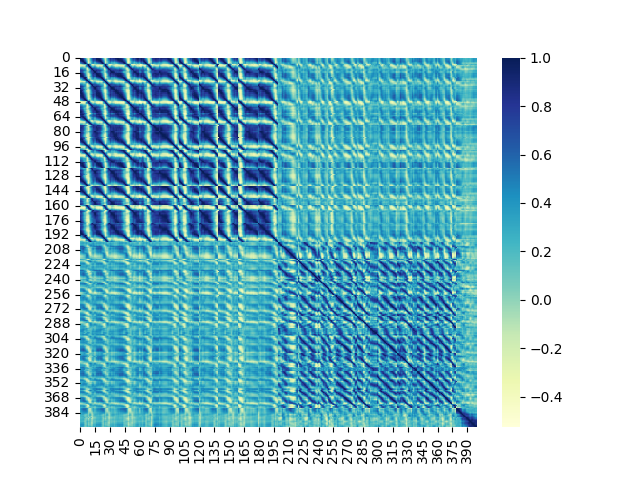} & 
    \includegraphics[scale=0.4]{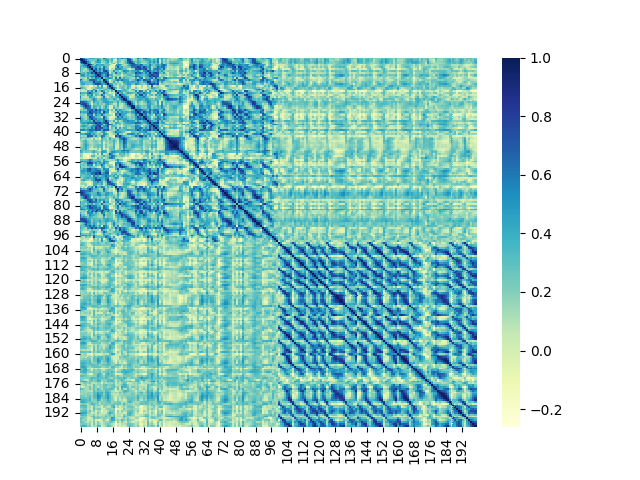} \\ 
    \footnotesize{(a) Cartpole-Swingup} & \footnotesize{(b)
    Cheetah-Run} & \footnotesize{(c) Walker-walk}
  \end{tabular}
    \caption{\footnotesize{Feature correlation heat-maps for three environments
    showing the correlation between the latent features obtained with
  CURL and CRC-RL models. The models are
trained for 49K environment steps and 200 latent features are used to
generate this plot.}}
\label{fig:corr_mat}
\end{figure*}

Another study is performed to validate our hypothesis that the
proposed CRC loss contribute new information resulting in learning new
feature representations which are distinct from those obtained using individual
losses. In this study, the correlation matrices between the latent
features obtained with the baseline CURL algorithm (that uses
contrastive loss) and that with
the proposed CRC-RL model (that uses CRC loss) is plotted as heat-maps
as shown in Figure \ref{fig:corr_mat}.  These matrices are generated
by collecting $200$ sample embeddings from both models trained with
49000 environment steps. Since, each image sample is encoded into a
$50\times 1$ feature vector, the features generated by the above two
methods are grouped into two feature matrices (say, $F_1$ and $F_2$)
of size $200\times 50$. The correlation between these two feature
matrices results in a $400x400$ matrix is visualized as a heat map in
the above figure where the darker regions shows higher correlation and
lighter regions show lower correlation. It is observed that
off-diagonal regions have lower correlation (lighter regions)
indicating that the two feature embeddings ($F1$ and $F2$ ) are very
distinct from each other. The diagonal regions are highly correlated
(darker regions) as they correspond to the features from the same
method. Another interesting finding of this study is that
these heat maps show increasingly complex patterns for difficult
environments such as 'Walker-walk' or 'Cheetah-Run' compared to simpler
environments such as 'Cartpole-Swingup'. These patterns evolve over time and
stabilize as the training  performance saturates. This is an
interesting insight that may provide clue to better understand the
relationship between features and action policies learned in an
end-to-end RL framework.

\subsection{Ablation Study} 
Three separate ablation studies are carried out to justify the design
choices made in this paper as described below.
\subsubsection{Usefulness of CRC loss function}
 First study validates the usefulness of the
proposed CRC loss function comprising of contrastive, reconstruction
and consistency losses. The outcome is shown in Figure
\ref{tab:loss_functions}. We start with the baseline CURL model
\cite{laskin2020curl} that uses contrastive loss to learn the feature
presentations. Then this model is trained with a combined loss
function of contrastive and reconstruction loss and finally, with the
CRC loss function comprising of contrastive, reconstruction and
consistency losses. The inclusion of these losses require modifying
the existing CURL model leading to the formation of the CRC-RL model
proposed in this paper. The figure shows that CRC-RL performs better
than the other two for the benchmark problems from the Deepmind
control suite (DMC), namely, `Cheetah-Run' or `Walker-walk'. These two
are comparatively difficult problems to solve compared to simpler
problems such as the `Cartpole-Swingup' problem which does not benefit
from the proposed CRC-RL model. This observation clearly establishes
the usefulness of the proposed approach. This becomes more evident in
the second ablation study discussed in the next section below.

\subsubsection{Relative weights of loss components in the CRC loss
function} In this study, the relative weights of three loss
components, namely, contrastive, reconstruction and consistency loss,
are varied and its effect on the validation performance is compared as
shown in Figure \ref{fig:loss_weights}. The weights are varied with
the constraint of forming a convex sum. In other words, $c_i>0,
i\in{1,2,3}$ and $\sum_ic_i = 1$. The figure shows that the individual
loss functions do not always provide the best performance. The best
performance is obtained by a combination of all the three losses.
Having equal weights ($c_1=c_2=c_3=0.33$) for all the three losses
have a regularizing effect on model performance in the sense that the
performance is traded for better generalizability. This is evident
from the fact that the validation performance curve with this
combination of weights lie somewhere in the middle of the all the
curves. The generalization capability of the proposed model is
demonstrated in the third ablation study discussed next.

\subsubsection{Generalization Capability of the CRC-RL model}
\begin{figure*}
\centering
\begin{tabular}{ccc}
\includegraphics[scale=0.08]{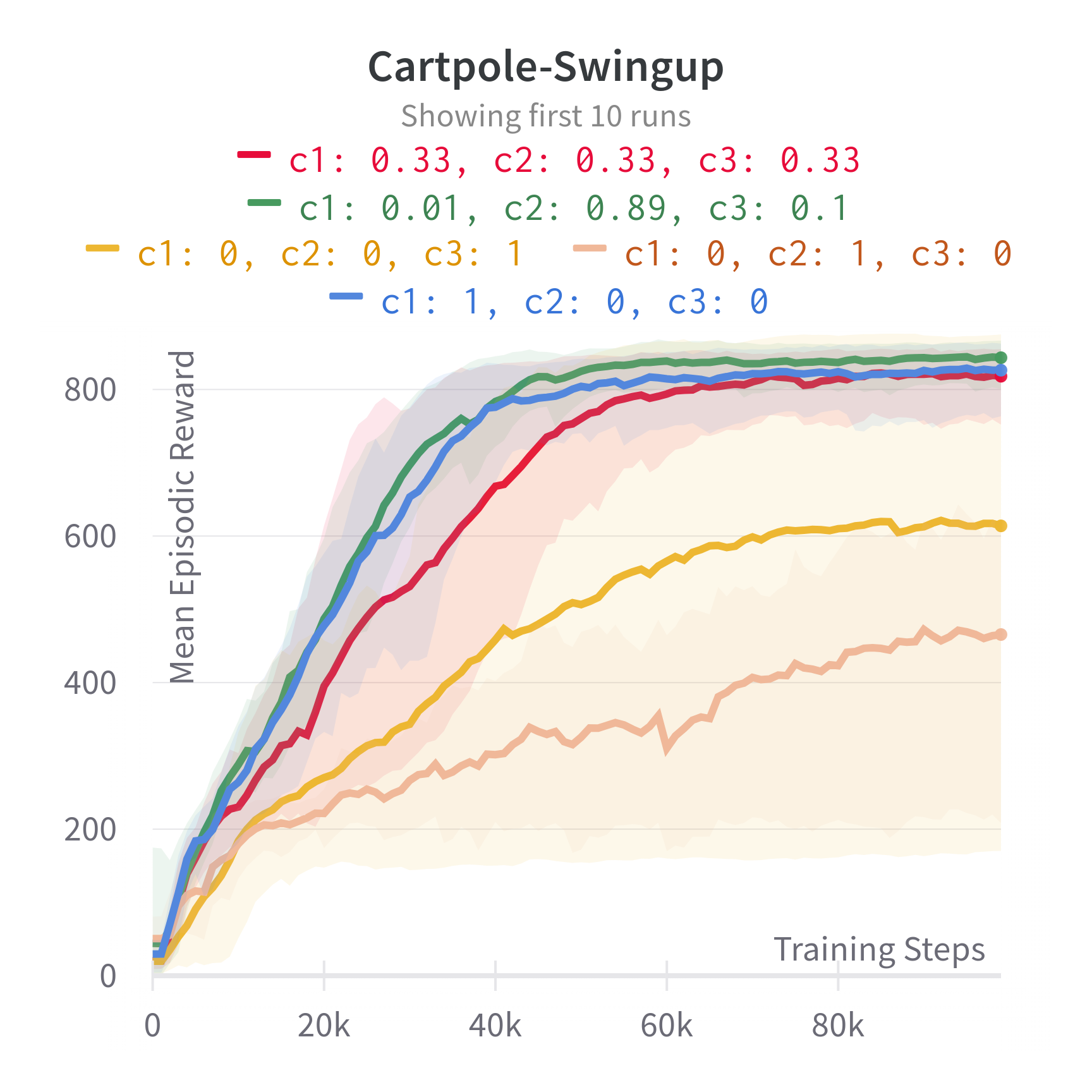}&
\includegraphics[scale=0.08]{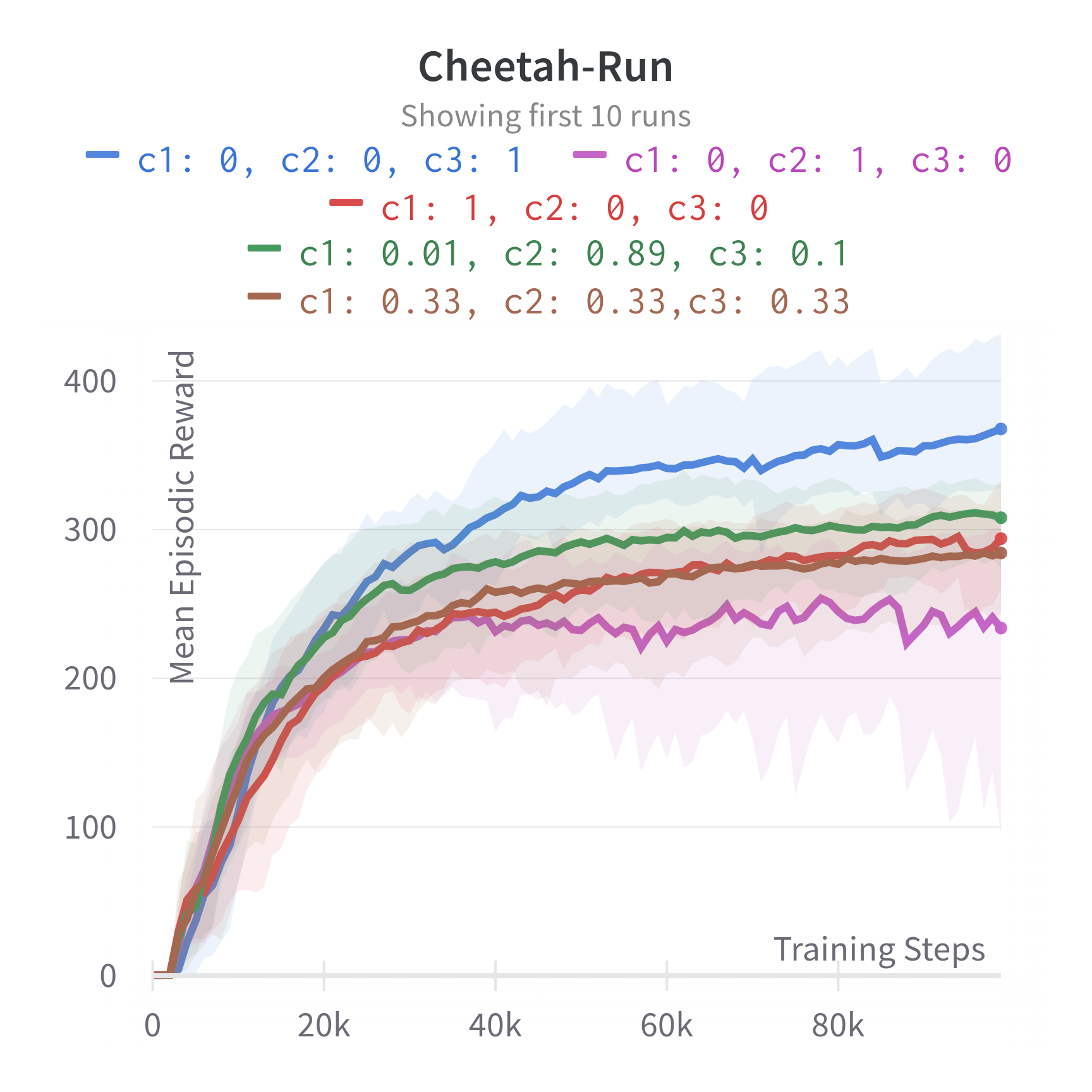} &
\includegraphics[scale=0.08]{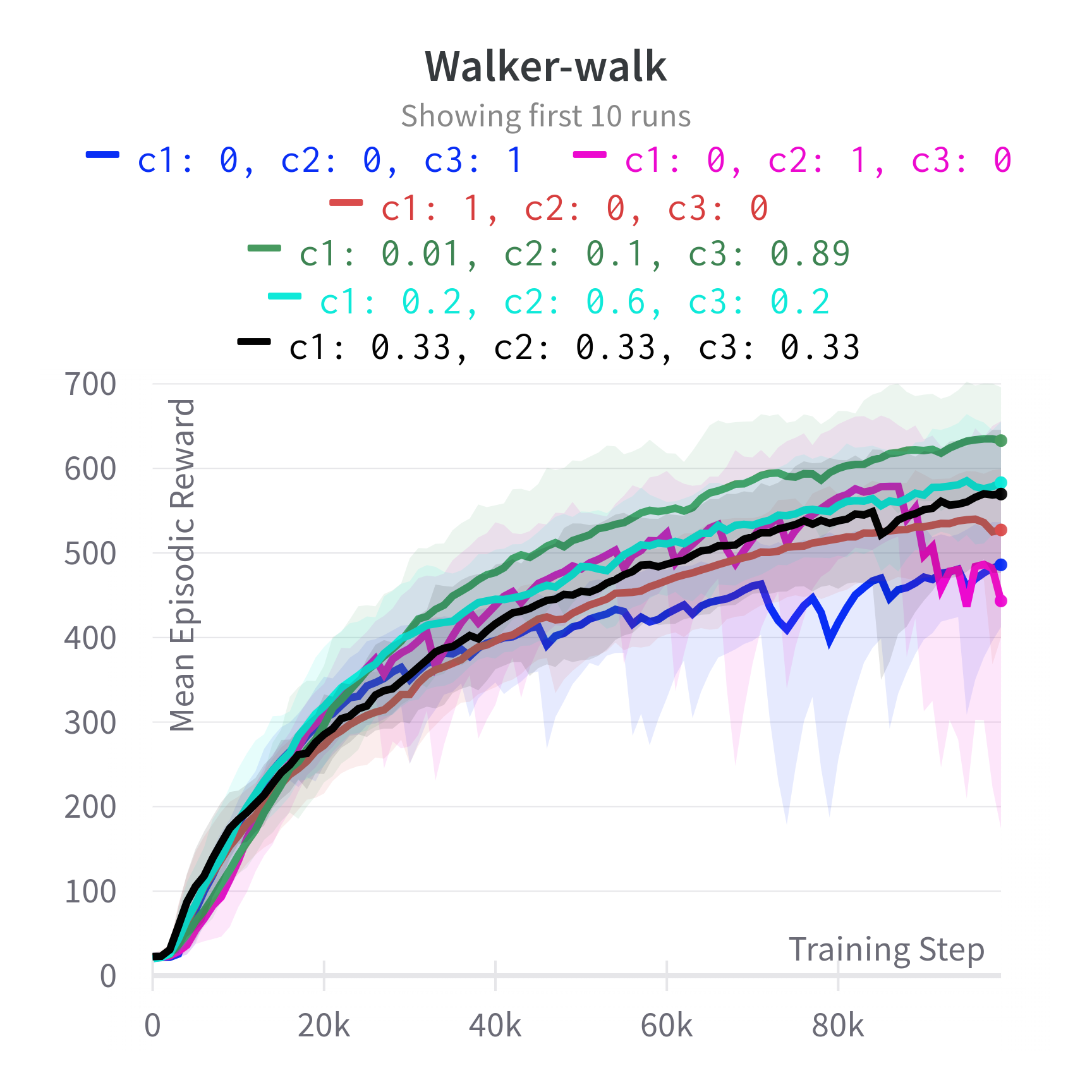}\\
\footnotesize{(a)} & \footnotesize{(b)} & \footnotesize{(c)}
\end{tabular}
\caption{\small{Effect of varying weighting parameters
for different loss functions on the Evaluation performance. $c_1$,
$c_2$ and $c_3$ are the weights to the contrastive loss, the reconstruction loss
and the consistency loss respectively in the CRC loss function. The
environments used are: (a) Cartpole-Swingup, (b) Cheetah-Run and (c)
Walker-walk. Varying these parameters have a regularizing effect on
the training performance. A smoothing factor of 0.5 is applied to the
plot.}} 
\label{fig:loss_weights}
\end{figure*}

\begin{figure*}
  \centering
  \begin{tabular}{cccc}
    \includegraphics[scale=0.06]{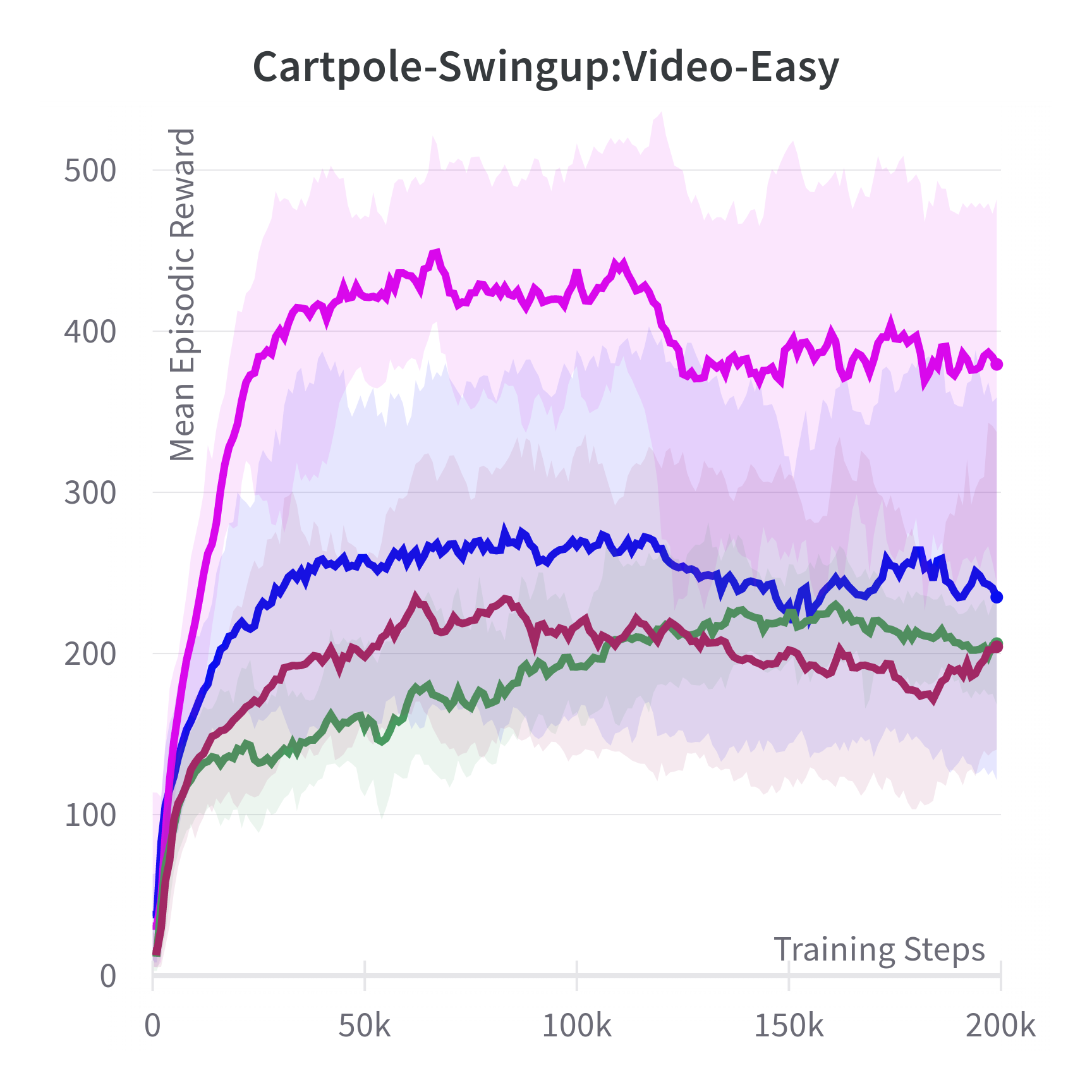} & 
    \includegraphics[scale=0.06]{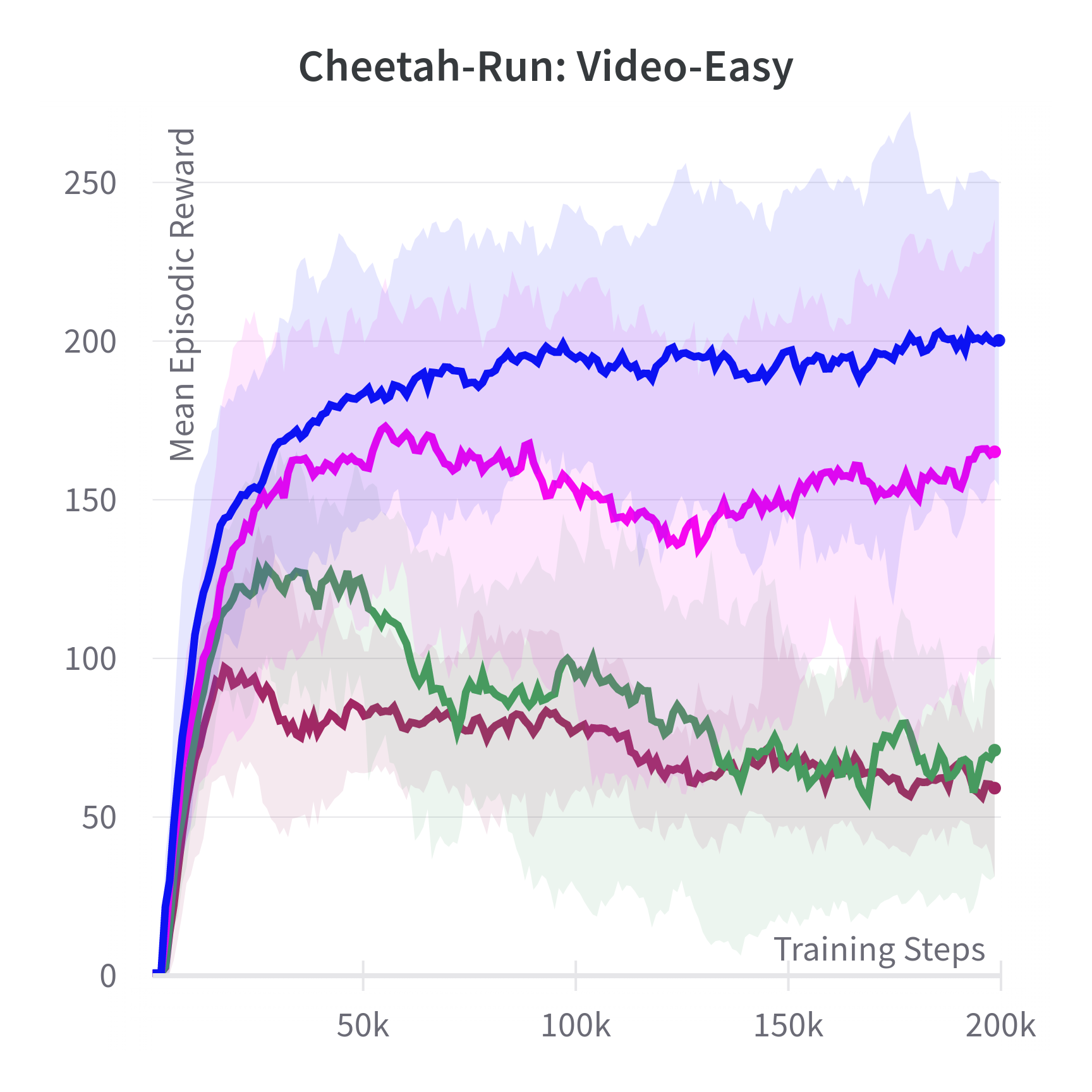} & 
    \includegraphics[scale=0.06]{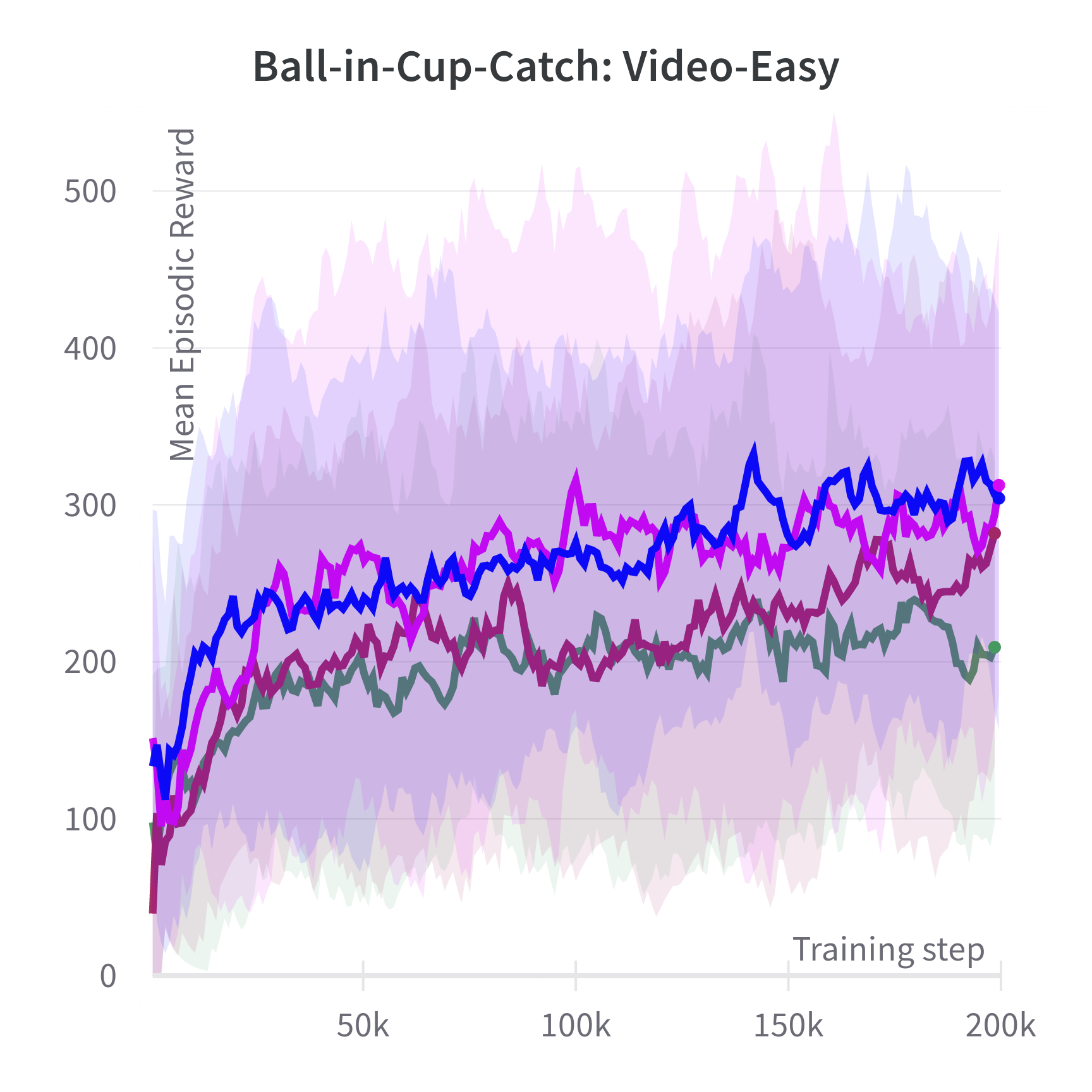} & 
    \includegraphics[scale=0.06]{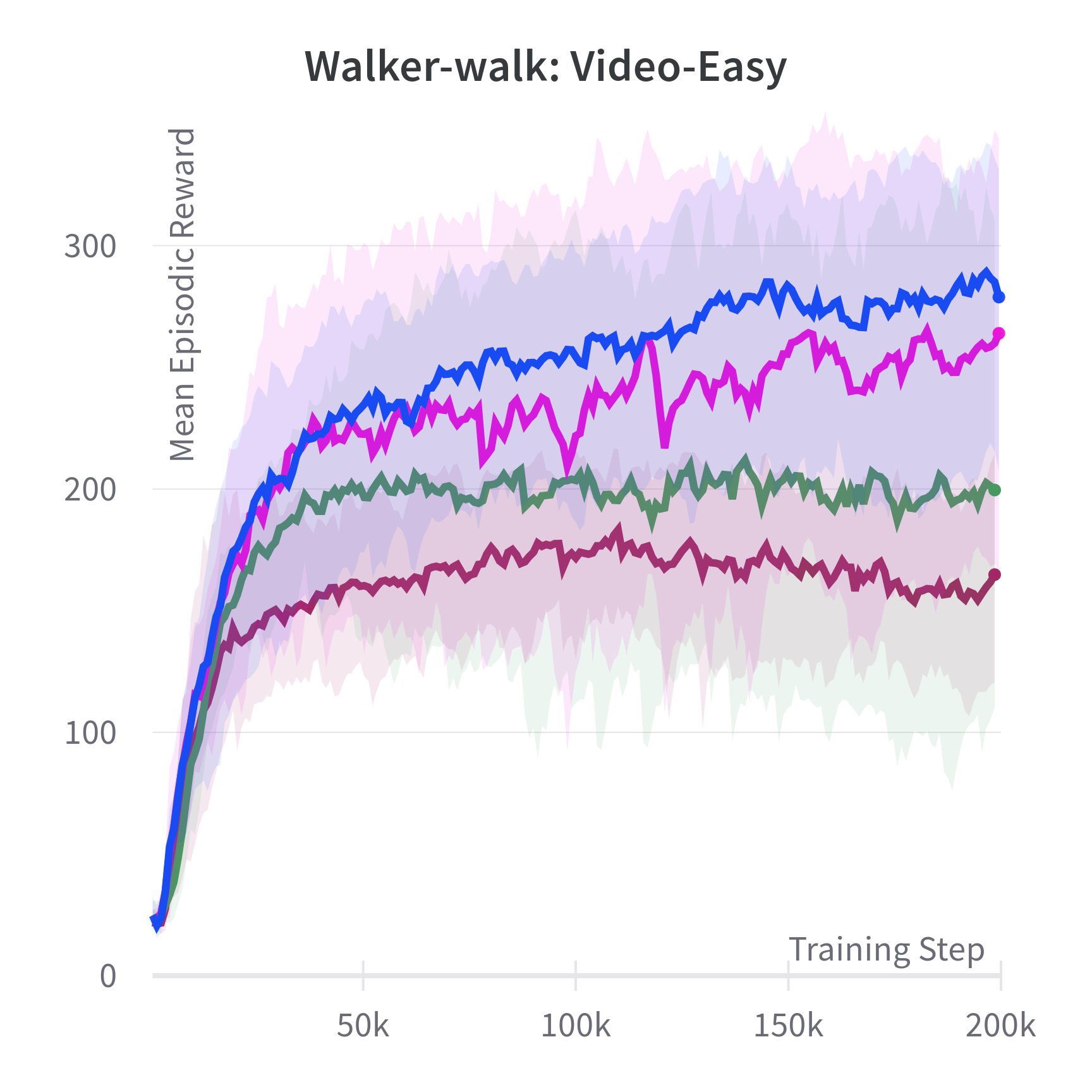} \\ 
    \includegraphics[scale=0.06]{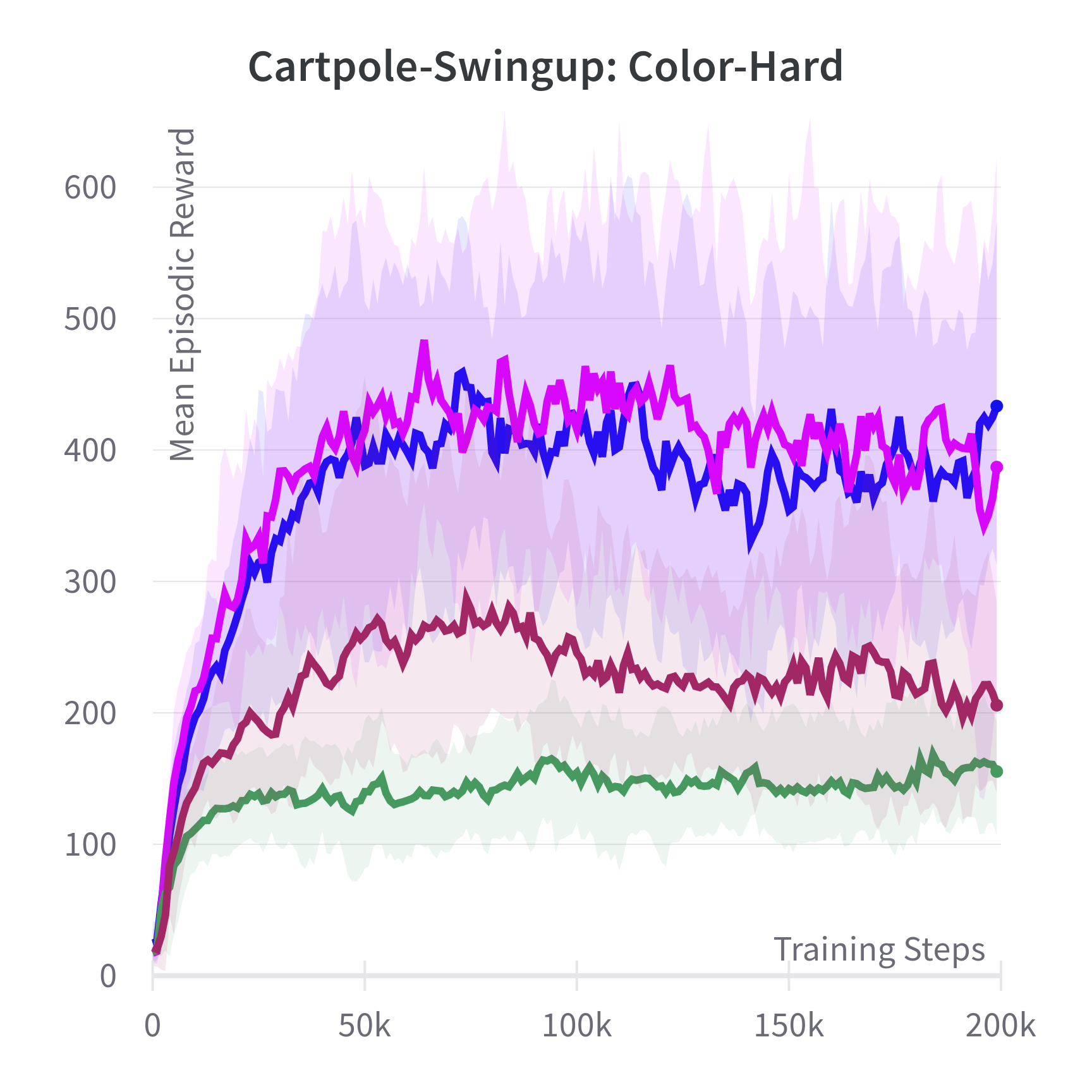} & 
    \includegraphics[scale=0.06]{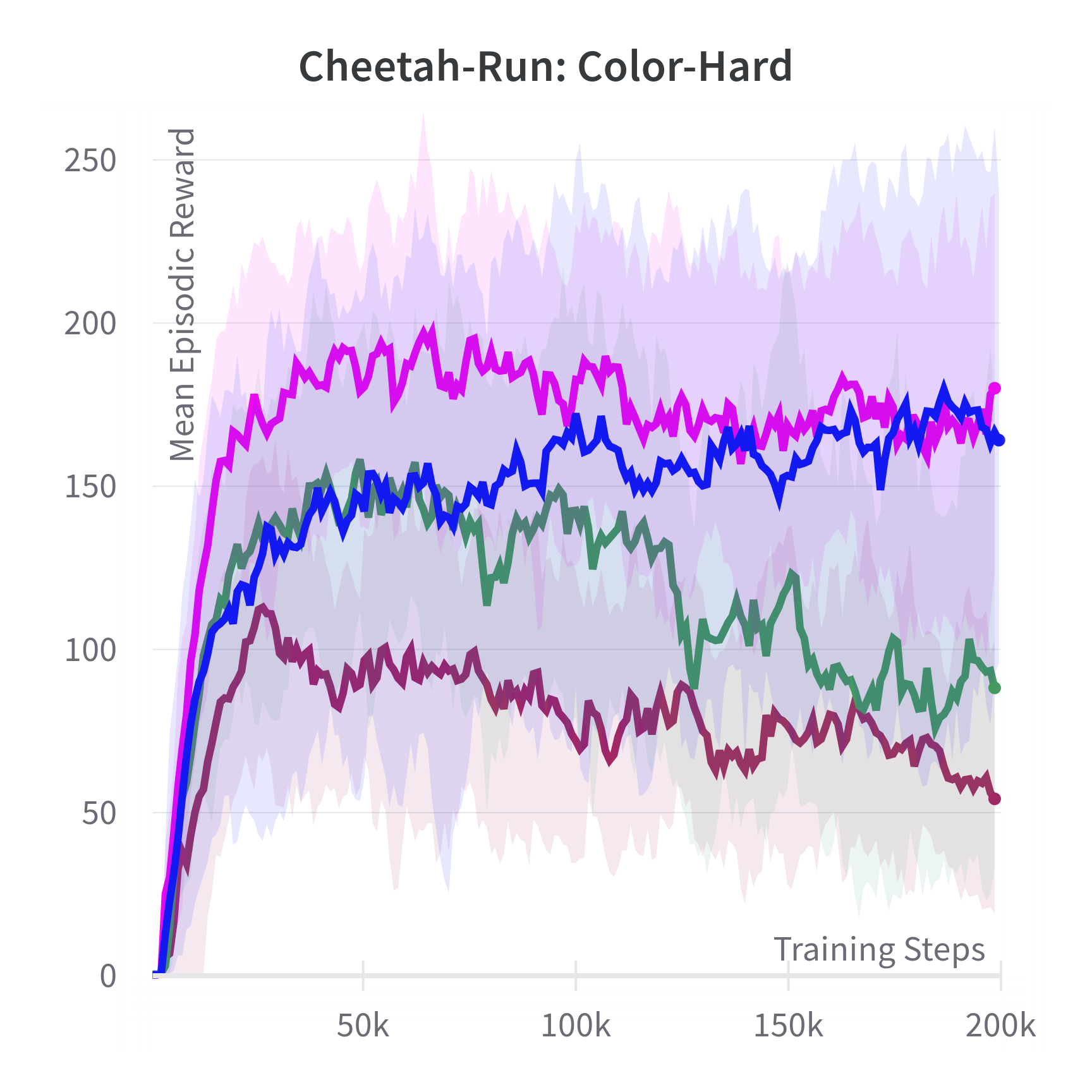} & 
    \includegraphics[scale=0.06]{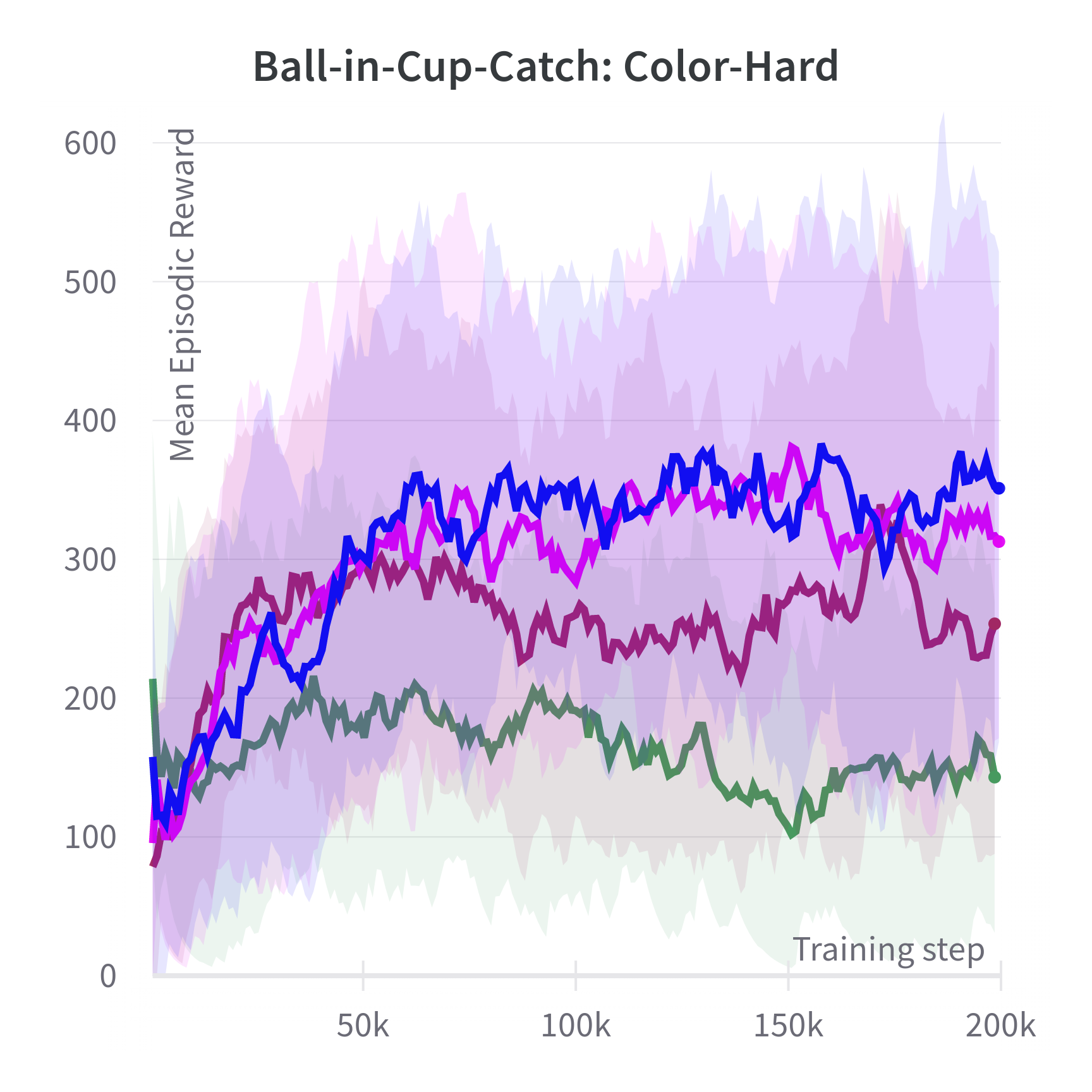} & 
    \includegraphics[scale=0.06]{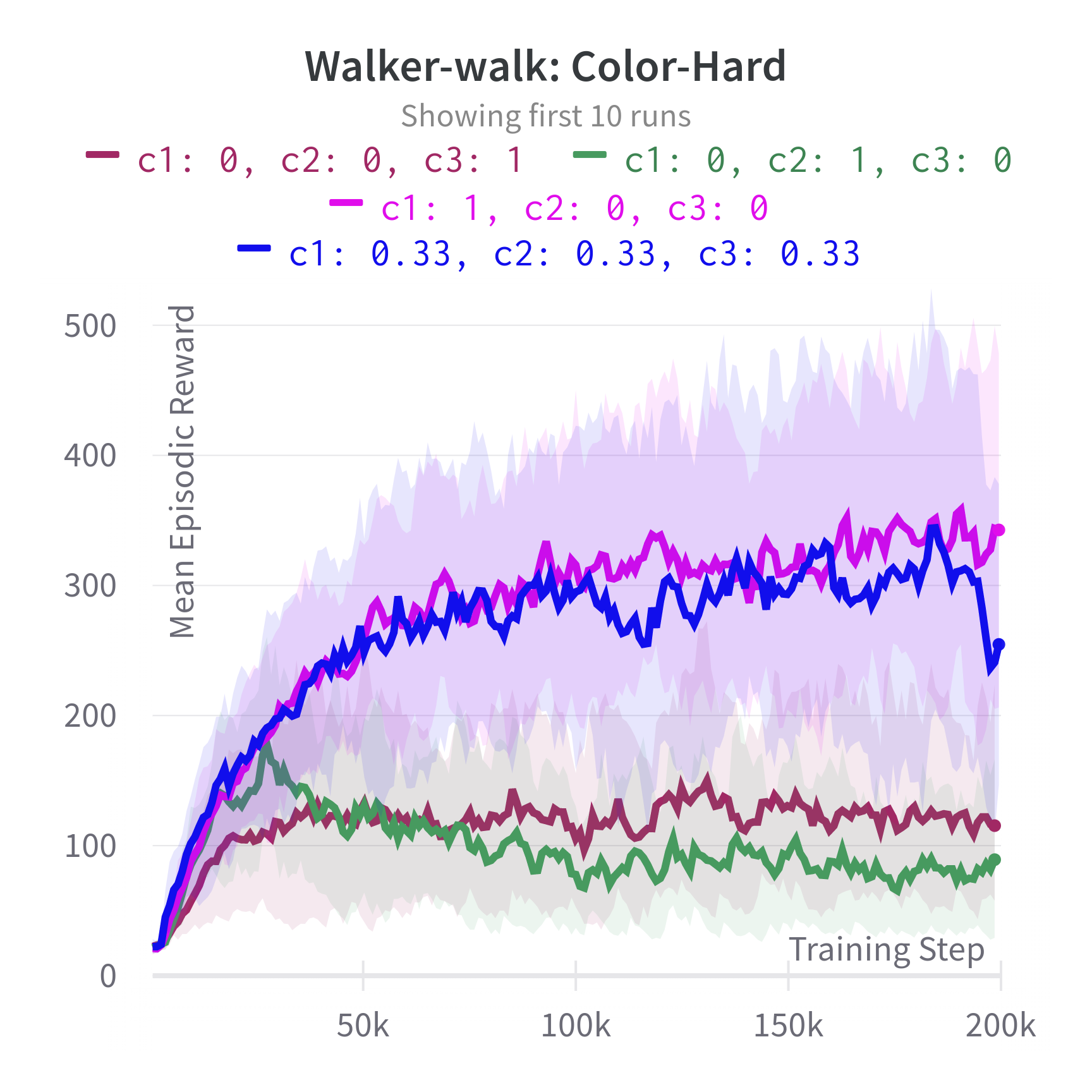} \\ 
\footnotesize{(a)} & \footnotesize{(b)} & \footnotesize{(c)} &
\footnotesize{(d)}
  \end{tabular}
  \caption{\small{Generalization capabilities of CRC-RL algorithm.
    $c_1$, $c_2$ and $c_3$ are the weights to the contrastive loss,
    the reconstruction loss and the consistency loss respectively in
    the CRC loss function. The environments used are: (a)
    Cartpole-Swingup, (b) Cheetah-Run and (c) Ball-in-Cup-Catch and (d) Walker-walk. The RL
    models  are trained on images with \emph{random-crop} augmentation
    and evaluated on images with \emph{Video-Easy} and
    \emph{Color-Hard} augmentations. Compared to
    the individual losses, CRC loss provide best or second-best
  evaluation performance for these new augmentations thereby
establishing the superior generalization capabilities of the CRC-RL
model. }}
  \label{fig:gen_effect}
\end{figure*}

In order to test the generalization capabilities of the proposed
CRC-RL model, another experiment is performed where the RL models are
trained on images augmented with \emph{random crop} effect and then
validated on images augmented with \emph{Video-Easy} and
\emph{Color-Hard} artifacts \cite{hansen2021generalization}. The
outcome is shown in Figure \ref{fig:gen_effect}.  Comparing with the
validation plots in Figure \ref{fig:loss_weights}, it is observed
that the overall performance has come down significantly due
to these complex augmentations which makes it difficult for the model
to generalize when trained only with images augmented with only
\emph{random crop} artefact. However, even in this case, the RL model
trained with CRC loss function provides the best or closer-to-the-best
evaluation performance compared to the models that use individual loss
components for training. This demonstrates the superior generalization
capability of the proposed CRC-RL model over the existing models that
use one of the above loss functions for training. It also corroborates
the earlier finding mentioned in the previous subsection that the
equal weights for individual loss components in the CRC loss function
have a regularizing effect on the model performance.

\section{Conclusions} \label{sec:conc} 
The paper addresses the problem of feature representation learning in
end-to-end reinforcement learning models with visual observations.
Specifically, a new loss function, called CRC loss, is proposed to
learn action-dependent features leading to superior performance in
learning optimal action policies. This loss function is a combination
of three different loss functions, namely, the image reconstruction
loss, the contrastive loss and the consistency loss. Through empirical
analysis including latent feature visualization, an attempt is made to
generate new insights that can better explain the relationship between
the features being learnt and the actions being taken by the RL agent.
The resulting architecture is shown to outperform the existing
state-of-the-art methods in solving the challenging DMC problems by a
significant margin thereby forming a new benchmark in this field. The
future work will involve carrying out more in-depth analysis and
evaluation of the individual loss components on the overall
performance as well as on the quality of features being learned.

{\small
	\bibliographystyle{unsrt}

	\bibliography{ref}

\begin{thebibliography}{10}

\bibitem{mnih2015human}
Volodymyr Mnih, Koray Kavukcuoglu, David Silver, Andrei~A Rusu, Joel Veness,
  Marc~G Bellemare, Alex Graves, Martin Riedmiller, Andreas~K Fidjeland, Georg
  Ostrovski, et~al.
\newblock Human-level control through deep reinforcement learning.
\newblock {\em nature}, 518(7540):529--533, 2015.

\bibitem{xue2022event}
Shan Xue, Biao Luo, Derong Liu, and Ying Gao.
\newblock Event-triggered integral reinforcement learning for nonzero-sum games
  with asymmetric input saturation.
\newblock {\em Neural Networks}, 152:212--223, 2022.

\bibitem{levine2016end}
Sergey Levine, Chelsea Finn, Trevor Darrell, and Pieter Abbeel.
\newblock End-to-end training of deep visuomotor policies.
\newblock {\em The Journal of Machine Learning Research}, 17(1):1334--1373,
  2016.

\bibitem{qureshi2018intrinsically}
Ahmed~Hussain Qureshi, Yutaka Nakamura, Yuichiro Yoshikawa, and Hiroshi
  Ishiguro.
\newblock Intrinsically motivated reinforcement learning for human--robot
  interaction in the real-world.
\newblock {\em Neural Networks}, 107:23--33, 2018.

\bibitem{wang2021modular}
Jiexin Wang, Stefan Elfwing, and Eiji Uchibe.
\newblock Modular deep reinforcement learning from reward and punishment for
  robot navigation.
\newblock {\em Neural Networks}, 135:115--126, 2021.

\bibitem{nakamura2007reinforcement}
Yutaka Nakamura, Takeshi Mori, Masa-aki Sato, and Shin Ishii.
\newblock Reinforcement learning for a biped robot based on a cpg-actor-critic
  method.
\newblock {\em Neural networks}, 20(6):723--735, 2007.

\bibitem{yang2018visual}
Wei Yang, Xiaolong Wang, Ali Farhadi, Abhinav Gupta, and Roozbeh Mottaghi.
\newblock Visual semantic navigation using scene priors.
\newblock {\em arXiv preprint arXiv:1810.06543}, 2018.

\bibitem{zhu2017target}
Yuke Zhu, Roozbeh Mottaghi, Eric Kolve, Joseph~J Lim, Abhinav Gupta,
  Li~Fei-Fei, and Ali Farhadi.
\newblock Target-driven visual navigation in indoor scenes using deep
  reinforcement learning.
\newblock In {\em 2017 IEEE international conference on robotics and automation
  (ICRA)}, pages 3357--3364. IEEE, 2017.

\bibitem{laskin2020reinforcement}
Misha Laskin, Kimin Lee, Adam Stooke, Lerrel Pinto, Pieter Abbeel, and Aravind
  Srinivas.
\newblock Reinforcement learning with augmented data.
\newblock {\em Advances in neural information processing systems},
  33:19884--19895, 2020.

\bibitem{laskin2020curl}
Michael Laskin, Aravind Srinivas, and Pieter Abbeel.
\newblock Curl: Contrastive unsupervised representations for reinforcement
  learning.
\newblock In {\em International Conference on Machine Learning}, pages
  5639--5650. PMLR, 2020.

\bibitem{zhuang2020comprehensive}
Fuzhen Zhuang, Zhiyuan Qi, Keyu Duan, Dongbo Xi, Yongchun Zhu, Hengshu Zhu, Hui
  Xiong, and Qing He.
\newblock A comprehensive survey on transfer learning.
\newblock {\em Proceedings of the IEEE}, 109(1):43--76, 2020.

\bibitem{islam2022transfer}
Tariqul Islam, Dm~Mehedi~Hasan Abid, Tanvir Rahman, Zahura Zaman, Kausar Mia,
  and Ramim Hossain.
\newblock Transfer learning in deep reinforcement learning.
\newblock In {\em Proceedings of Seventh International Congress on Information
  and Communication Technology: ICICT 2022, London, Volume 1}, pages 145--153.
  Springer, 2022.

\bibitem{jaafra2018review}
Yesmina Jaafra, Jean~Luc Laurent, Aline Deruyver, and Mohamed~Saber Naceur.
\newblock A review of meta-reinforcement learning for deep neural networks
  architecture search.
\newblock {\em arXiv preprint arXiv:1812.07995}, 2018.

\bibitem{hospedales2021meta}
Timothy Hospedales, Antreas Antoniou, Paul Micaelli, and Amos Storkey.
\newblock Meta-learning in neural networks: A survey.
\newblock {\em IEEE transactions on pattern analysis and machine intelligence},
  44(9):5149--5169, 2021.

\bibitem{settles2009active}
Burr Settles.
\newblock Active learning literature survey.
\newblock 2009.

\bibitem{bengio2013representation}
Yoshua Bengio, Aaron Courville, and Pascal Vincent.
\newblock Representation learning: A review and new perspectives.
\newblock {\em IEEE transactions on pattern analysis and machine intelligence},
  35(8):1798--1828, 2013.

\bibitem{yarats2021improving}
Denis Yarats, Amy Zhang, Ilya Kostrikov, Brandon Amos, Joelle Pineau, and Rob
  Fergus.
\newblock Improving sample efficiency in model-free reinforcement learning from
  images.
\newblock In {\em Proceedings of the AAAI Conference on Artificial
  Intelligence}, volume~35, pages 10674--10681, 2021.

\bibitem{ericsson2022self}
Linus Ericsson, Henry Gouk, Chen~Change Loy, and Timothy~M Hospedales.
\newblock Self-supervised representation learning: Introduction, advances, and
  challenges.
\newblock {\em IEEE Signal Processing Magazine}, 39(3):42--62, 2022.

\bibitem{bank2020autoencoders}
Dor Bank, Noam Koenigstein, and Raja Giryes.
\newblock Autoencoders.
\newblock {\em arXiv preprint arXiv:2003.05991}, 2020.

\bibitem{lin2017marta}
Daoyu Lin, Kun Fu, Yang Wang, Guangluan Xu, and Xian Sun.
\newblock Marta gans: Unsupervised representation learning for remote sensing
  image classification.
\newblock {\em IEEE Geoscience and Remote Sensing Letters}, 14(11):2092--2096,
  2017.

\bibitem{peng2019cm}
Yuxin Peng and Jinwei Qi.
\newblock Cm-gans: Cross-modal generative adversarial networks for common
  representation learning.
\newblock {\em ACM Transactions on Multimedia Computing, Communications, and
  Applications (TOMM)}, 15(1):1--24, 2019.

\bibitem{chen2020simple}
Ting Chen, Simon Kornblith, Mohammad Norouzi, and Geoffrey Hinton.
\newblock A simple framework for contrastive learning of visual
  representations.
\newblock In {\em International conference on machine learning}, pages
  1597--1607. PMLR, 2020.

\bibitem{hansen2021generalization}
Nicklas Hansen and Xiaolong Wang.
\newblock Generalization in reinforcement learning by soft data augmentation.
\newblock In {\em 2021 IEEE International Conference on Robotics and Automation
  (ICRA)}, pages 13611--13617. IEEE, 2021.

\bibitem{finn2015learning}
Chelsea Finn, Xin~Yu Tan, Yan Duan, Trevor Darrell, Sergey Levine, and Pieter
  Abbeel.
\newblock Learning visual feature spaces for robotic manipulation with deep
  spatial autoencoders.
\newblock {\em arXiv preprint arXiv:1509.06113}, 25:2, 2015.

\bibitem{stooke2021decoupling}
Adam Stooke, Kimin Lee, Pieter Abbeel, and Michael Laskin.
\newblock Decoupling representation learning from reinforcement learning.
\newblock In {\em International Conference on Machine Learning}, pages
  9870--9879. PMLR, 2021.

\bibitem{jaderberg2016reinforcement}
Max Jaderberg, Volodymyr Mnih, Wojciech~Marian Czarnecki, Tom Schaul, Joel~Z
  Leibo, David Silver, and Koray Kavukcuoglu.
\newblock Reinforcement learning with unsupervised auxiliary tasks.
\newblock {\em arXiv preprint arXiv:1611.05397}, 2016.

\bibitem{xenou2018deep}
Konstantia Xenou, Georgios Chalkiadakis, and Stergos Afantenos.
\newblock Deep reinforcement learning in strategic board game environments.
\newblock In {\em European Conference on Multi-Agent Systems}, pages 233--248.
  Springer, 2018.

\bibitem{neyshabur2020observational}
Behnam Neyshabur, Stephen Tu, Xingyou Song, Yiding Jiang, and Yilun Du.
\newblock Observational overfitting in reinforcement learning.
\newblock 2020.

\bibitem{tassa2018deepmind}
Yuval Tassa, Yotam Doron, Alistair Muldal, Tom Erez, Yazhe Li, Diego de~Las
  Casas, David Budden, Abbas Abdolmaleki, Josh Merel, Andrew Lefrancq, et~al.
\newblock Deepmind control suite.
\newblock {\em arXiv preprint arXiv:1801.00690}, 2018.

\bibitem{sutton2018reinforcement}
Richard~S Sutton and Andrew~G Barto.
\newblock {\em Reinforcement learning: An introduction}.
\newblock MIT press, 2018.

\bibitem{barto1995reinforcement}
Andrew~G Barto.
\newblock Reinforcement learning and dynamic programming.
\newblock In {\em Analysis, Design and Evaluation of Man--Machine Systems
  1995}, pages 407--412. Elsevier, 1995.

\bibitem{van2016deep}
Hado Van~Hasselt, Arthur Guez, and David Silver.
\newblock Deep reinforcement learning with double q-learning.
\newblock In {\em Proceedings of the AAAI Conference on Artificial
  Intelligence}, volume~30, 2016.

\bibitem{sewak2019deep}
Mohit Sewak.
\newblock Deep q network (dqn), double dqn, and dueling dqn.
\newblock In {\em Deep Reinforcement Learning}, pages 95--108. Springer, 2019.

\bibitem{mnih2013playing}
Volodymyr Mnih, Koray Kavukcuoglu, David Silver, Alex Graves, Ioannis
  Antonoglou, Daan Wierstra, and Martin Riedmiller.
\newblock Playing atari with deep reinforcement learning.
\newblock {\em arXiv preprint arXiv:1312.5602}, 2013.

\bibitem{holcomb2018overview}
Sean~D Holcomb, William~K Porter, Shaun~V Ault, Guifen Mao, and Jin Wang.
\newblock Overview on deepmind and its alphago zero ai.
\newblock In {\em Proceedings of the 2018 international conference on big data
  and education}, pages 67--71, 2018.

\bibitem{arulkumaran2017deep}
Kai Arulkumaran, Marc~Peter Deisenroth, Miles Brundage, and Anil~Anthony
  Bharath.
\newblock Deep reinforcement learning: A brief survey.
\newblock {\em IEEE Signal Processing Magazine}, 34(6):26--38, 2017.

\bibitem{lillicrap2015continuous}
Timothy~P Lillicrap, Jonathan~J Hunt, Alexander Pritzel, Nicolas Heess, Tom
  Erez, Yuval Tassa, David Silver, and Daan Wierstra.
\newblock Continuous control with deep reinforcement learning.
\newblock {\em arXiv preprint arXiv:1509.02971}, 2015.

\bibitem{duan2016benchmarking}
Yan Duan, Xi~Chen, Rein Houthooft, John Schulman, and Pieter Abbeel.
\newblock Benchmarking deep reinforcement learning for continuous control.
\newblock In {\em International conference on machine learning}, pages
  1329--1338. PMLR, 2016.

\bibitem{gu2017deep}
Shixiang Gu, Ethan Holly, Timothy Lillicrap, and Sergey Levine.
\newblock Deep reinforcement learning for robotic manipulation with
  asynchronous off-policy updates.
\newblock In {\em 2017 IEEE international conference on robotics and automation
  (ICRA)}, pages 3389--3396. IEEE, 2017.

\bibitem{nguyen2019review}
Hai Nguyen and Hung La.
\newblock Review of deep reinforcement learning for robot manipulation.
\newblock In {\em 2019 Third IEEE International Conference on Robotic Computing
  (IRC)}, pages 590--595. IEEE, 2019.

\bibitem{quillen2018deep}
Deirdre Quillen, Eric Jang, Ofir Nachum, Chelsea Finn, Julian Ibarz, and Sergey
  Levine.
\newblock Deep reinforcement learning for vision-based robotic grasping: A
  simulated comparative evaluation of off-policy methods.
\newblock In {\em 2018 IEEE International Conference on Robotics and Automation
  (ICRA)}, pages 6284--6291. IEEE, 2018.

\bibitem{joshi2020robotic}
Shirin Joshi, Sulabh Kumra, and Ferat Sahin.
\newblock Robotic grasping using deep reinforcement learning.
\newblock In {\em 2020 IEEE 16th International Conference on Automation Science
  and Engineering (CASE)}, pages 1461--1466. IEEE, 2020.

\bibitem{yue2019experimental}
Pengyu Yue, Jing Xin, Huan Zhao, Ding Liu, Mao Shan, and Jian Zhang.
\newblock Experimental research on deep reinforcement learning in autonomous
  navigation of mobile robot.
\newblock In {\em 2019 14th IEEE Conference on Industrial Electronics and
  Applications (ICIEA)}, pages 1612--1616. IEEE, 2019.

\bibitem{mnih2016asynchronous}
Volodymyr Mnih, Adria~Puigdomenech Badia, Mehdi Mirza, Alex Graves, Timothy
  Lillicrap, Tim Harley, David Silver, and Koray Kavukcuoglu.
\newblock Asynchronous methods for deep reinforcement learning.
\newblock In {\em International conference on machine learning}, pages
  1928--1937. PMLR, 2016.

\bibitem{haarnoja2018soft}
Tuomas Haarnoja, Aurick Zhou, Kristian Hartikainen, George Tucker, Sehoon Ha,
  Jie Tan, Vikash Kumar, Henry Zhu, Abhishek Gupta, Pieter Abbeel, et~al.
\newblock Soft actor-critic algorithms and applications.
\newblock {\em arXiv preprint arXiv:1812.05905}, 2018.

\bibitem{schulman2017proximal}
John Schulman, Filip Wolski, Prafulla Dhariwal, Alec Radford, and Oleg Klimov.
\newblock Proximal policy optimization algorithms.
\newblock {\em arXiv preprint arXiv:1707.06347}, 2017.

\bibitem{botteghi2022unsupervised}
Nicol{\`o} Botteghi, Mannes Poel, and Christoph Brune.
\newblock Unsupervised representation learning in deep reinforcement learning:
  A review.
\newblock {\em arXiv preprint arXiv:2208.14226}, 2022.

\bibitem{pinaya2020autoencoders}
Walter Hugo~Lopez Pinaya, Sandra Vieira, Rafael Garcia-Dias, and Andrea
  Mechelli.
\newblock Autoencoders.
\newblock In {\em Machine learning}, pages 193--208. Elsevier, 2020.

\bibitem{raffin2018s}
Antonin Raffin, Ashley Hill, Ren{\'e} Traor{\'e}, Timoth{\'e}e Lesort, Natalia
  D{\'\i}az-Rodr{\'\i}guez, and David Filliat.
\newblock S-rl toolbox: Environments, datasets and evaluation metrics for state
  representation learning.
\newblock {\em arXiv preprint arXiv:1809.09369}, 2018.

\bibitem{lange2010deep}
Sascha Lange and Martin Riedmiller.
\newblock Deep auto-encoder neural networks in reinforcement learning.
\newblock In {\em The 2010 international joint conference on neural networks
  (IJCNN)}, pages 1--8. IEEE, 2010.

\bibitem{kolesnikov2019revisiting}
Alexander Kolesnikov, Xiaohua Zhai, and Lucas Beyer.
\newblock Revisiting self-supervised visual representation learning.
\newblock In {\em Proceedings of the IEEE/CVF conference on computer vision and
  pattern recognition}, pages 1920--1929, 2019.

\bibitem{raileanu2020automatic}
Roberta Raileanu, Max Goldstein, Denis Yarats, Ilya Kostrikov, and Rob Fergus.
\newblock Automatic data augmentation for generalization in deep reinforcement
  learning.
\newblock {\em arXiv preprint arXiv:2006.12862}, 2020.

\bibitem{shang2021reinforcement}
Wenling Shang, Xiaofei Wang, Aravind Srinivas, Aravind Rajeswaran, Yang Gao,
  Pieter Abbeel, and Misha Laskin.
\newblock Reinforcement learning with latent flow.
\newblock {\em Advances in Neural Information Processing Systems},
  34:22171--22183, 2021.

\bibitem{henaff2020data}
Olivier Henaff.
\newblock Data-efficient image recognition with contrastive predictive coding.
\newblock In {\em International conference on machine learning}, pages
  4182--4192. PMLR, 2020.

\bibitem{oord2018representation}
Aaron van~den Oord, Yazhe Li, and Oriol Vinyals.
\newblock Representation learning with contrastive predictive coding.
\newblock {\em arXiv preprint arXiv:1807.03748}, 2018.

\bibitem{ghosh2019variational}
Partha Ghosh, Mehdi~SM Sajjadi, Antonio Vergari, Michael Black, and Bernhard
  Sch{\"o}lkopf.
\newblock From variational to deterministic autoencoders.
\newblock {\em arXiv preprint arXiv:1903.12436}, 2019.

\bibitem{paszke2019pytorch}
Adam Paszke, Sam Gross, Francisco Massa, Adam Lerer, James Bradbury, Gregory
  Chanan, Trevor Killeen, Zeming Lin, Natalia Gimelshein, Luca Antiga, et~al.
\newblock Pytorch: An imperative style, high-performance deep learning library.
\newblock {\em Advances in neural information processing systems}, 32, 2019.

\bibitem{yarats2020image}
Denis Yarats, Ilya Kostrikov, and Rob Fergus.
\newblock Image augmentation is all you need: Regularizing deep reinforcement
  learning from pixels.
\newblock In {\em International Conference on Learning Representations}, 2020.

\bibitem{hafner2019learning}
Danijar Hafner, Timothy Lillicrap, Ian Fischer, Ruben Villegas, David Ha,
  Honglak Lee, and James Davidson.
\newblock Learning latent dynamics for planning from pixels.
\newblock In {\em International conference on machine learning}, pages
  2555--2565. PMLR, 2019.

\bibitem{hafner2019dream}
Danijar Hafner, Timothy Lillicrap, Jimmy Ba, and Mohammad Norouzi.
\newblock Dream to control: Learning behaviors by latent imagination.
\newblock {\em arXiv preprint arXiv:1912.01603}, 2019.

\end{thebibliography}
}

\end{document}